\documentclass{article}

\usepackage[final]{corl_2021} 

\usepackage{times}

\usepackage{minitoc}
\usepackage[numbers]{natbib}
\usepackage{multicol}
\usepackage{graphicx}
\usepackage{wrapfig,lipsum,booktabs}
\usepackage[font=footnotesize,labelfont=bf]{caption}
\usepackage{subcaption}
\usepackage{bbm}
\usepackage{amsmath}
\usepackage{booktabs}
\usepackage{diagbox}
\usepackage{multirow}

\usepackage{array}
\usepackage{graphicx}
\usepackage[utf8]{inputenc}
\usepackage{amsmath, amsthm, amssymb}
\usepackage{dsfont}
\usepackage{url}
\usepackage[ruled,vlined,linesnumbered]{algorithm2e}

\newcolumntype{P}[1]{>{\centering\arraybackslash}p{#1}}

\newcommand{\algo}{VCD}
\SetCommentSty{mycommfont}

\title{Learning Visible Connectivity Dynamics for Cloth Smoothing}

\author{
  Xingyu Lin$^*$\\
  Robotics Institute\\
  Carnegie Mellon University \\
  \texttt{xlin3@andrew.cmu.edu} \\
   \And
    Yufei Wang$^*$\\
  Robotics Institute\\
  Carnegie Mellon University \\
  \texttt{yufeiw2@andrew.cmu.edu} \\
   \And
  Zixuan Huang\\
    Robotics Institute\\
  Carnegie Mellon University \\
  \texttt{zixuanhu@andrew.cmu.edu} \\
   \And
    David Held\\
      Robotics Institute\\
  Carnegie Mellon University \\
  \texttt{dheld@andrew.cmu.edu} \\
}

\begin{document}
\maketitle

{\let\thefootnote\relax\footnotetext{* Equal contribution, order by dice rolling.}}
\doparttoc 
\faketableofcontents 

\part{} 
\vspace{-0.8in}

\begin{abstract}
Robotic manipulation of cloth remains challenging due to the complex dynamics of cloth, lack of a low-dimensional state representation, and self-occlusions. In contrast to previous model-based approaches that learn a pixel-based dynamics model or a compressed latent vector dynamics, we propose to learn a particle-based dynamics model from a partial point cloud observation. To overcome the challenges of partial observability, we infer which visible points are connected on the underlying cloth mesh.  We then learn a dynamics model over this visible connectivity graph. Compared to previous learning-based approaches, our model poses strong inductive bias with its particle based representation for learning the underlying cloth physics; it can generalize to cloths with novel shapes; it is invariant to visual features; and the predictions can be more easily visualized. We show that our method greatly outperforms previous state-of-the-art model-based and model-free reinforcement learning methods in simulation. Furthermore, we demonstrate zero-shot sim-to-real transfer where we deploy the model trained in simulation on a Franka arm and show that the model can successfully smooth cloths of different materials, geometries and colors from crumpled configurations. Videos can be found in the supplement and on our anonymous project website.\footnote{\url{https://sites.google.com/view/vcd-cloth}}
\end{abstract}
\keywords{Deformable Object Manipulation,  Dynamics Modeling}

\section{Introduction}
Robotic manipulation of cloth has wide applications across both industrial and domestic tasks such as laundry folding and bed making. However, cloth manipulation remains challenging for robotics due to the complex cloth dynamics.  Further, like most deformable objects, cloth cannot be easily described by low-dimensional state representations when placed in arbitrary configurations. Self-occlusions make state estimation especially difficult when the cloth is crumpled.


One approach to cloth manipulation explored by previous work, which we also adopt, is to learn a cloth dynamics model and then use the model for planning to determine the robot actions.  However, given that a crumpled cloth has many self-occlusions and complex dynamics, it is unclear how to choose the appropriate state representation. One possible state representation is to use a mesh model of the entire cloth~\cite{jimenez2020perception}. However, fitting a full mesh model to an arbitrary crumpled cloth configuration is difficult. Recent work have approached fabric manipulation by either compressing the cloth representation into a fixed-size latent vector~\cite{yan2020learning,wu2019learning,matas2018sim} or directly learning a visual dynamics model in pixel space~\cite{fabric_vsf_2020}. However, these representations do not enforce any inductive bias of the cloth physics, leading to suboptimal performance and generalization.





In contrast to a pixel-based or latent dynamics model, particle-based models have recently been shown to be able to learn dynamics for fluid and plastics~\cite{li2018learning, sanchez2020learning,pfaff2020learning}. A particle-based dynamics representation has the following benefits: first, it captures the inductive bias of the underlying physics, since real-world objects are composed of underlying atoms that can be modeled on the micro-level by particles.
Second, we can incorporate inductive bias by directly applying the effect of the robot gripper on the particle being grasped (though the effect on the other particles must still be inferred). 
Last, particle-based models are invariant to visual features such as object colors or patterns.  
As such, in this paper we aim to learn a particle-based dynamics model for cloth. However, the challenges in applying the particle-based model to cloth are that we cannot directly observe the underlying particles composing the cloth nor their mesh connections. The problem is made even more challenging due to the partial observability of the cloth from self-occlusions when it is in a crumpled configuration.  


    
\begin{figure*}[t]
    \centering
    \includegraphics[width=\textwidth]{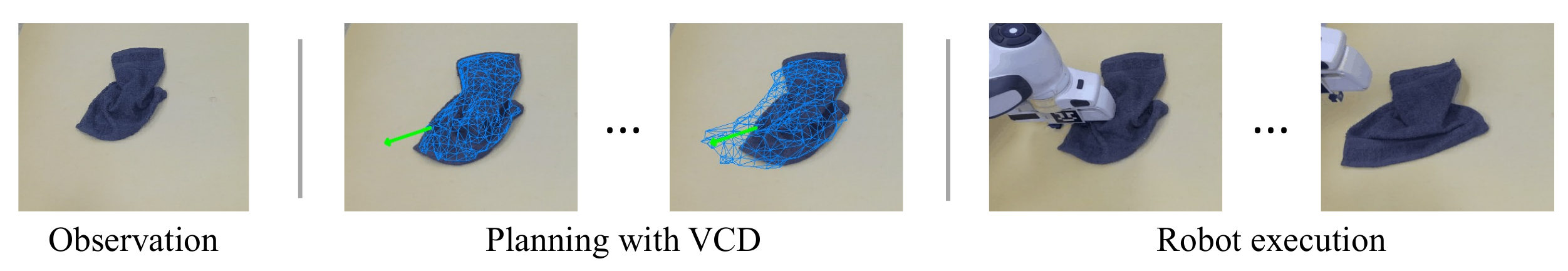}
    \caption{Cloth smoothing by planning using a dynamics model with a visible connectivity graph.}
    \label{fig:pull}
    \vspace{-3mm}
\end{figure*}

Our insight into this problem is that, rather than fitting a mesh model to the observation, we should
learn the \emph{visible connectivity dynamics (VCD):} a dynamics model based on the connectivity structure of the visible portion of the cloth.  To do so, we first learn to estimate the \emph{visible connectivity graph}: 
we estimate which points in the point cloud observation are connected in the underlying cloth mesh  (see  Figure~\ref{fig:pull}).  Estimating the mesh connectivity of the observation is a simplification of the problem of fitting a single full mesh model of the entire cloth to the observation; however, it is significantly easier to learn, since we do not need to find a globally consistent explanation of the observation which requires reasoning about occlusion; to estimate the mesh connectivity of the observation, we only need to consider the visible local cloth structure. While the graph is constructed only based on the visible points, we show that the dynamics model can be trained to be robust to partial observation.


In this work, we focus on the task of smoothing a piece of cloth from a crumpled configuration.
We propose a method that infers the observable particles and their connections from the point cloud, learns a visible connectivity dynamics model for the observable portion of the cloth, and uses it for planning to smooth the cloth. We show that for smoothing, planning with a visible connectivity dynamics model greatly outperforms state-of-the-art model-based and model-free reinforcement learning methods that use a fixed-size latent vector representation or learn a pixel-based visual dynamics model.  We then demonstrate zero-shot sim-to-real transfer where we deploy the model trained in simulation on a Franka arm and show that the learned model can successfully smooth cloths of different materials, geometries, and colors from crumpled configurations.

\section{Related Work}
    
\noindent \textbf{Vision-based Cloth Manipulation: }
Some papers on cloth manipulation assume that the cloth is already lying flat on the table~\cite{miller2011parametrized,stria2014polygonal,stria2014garment}. If the cloth starts in an unknown configuration, then one approach is to perform a sequence of actions that are designed to move the cloth into a set of known configurations from which perception can be performed more easily~\cite{cusumano2011bringing,maitin2010cloth,triantafyllou2011vision}. 
For example, the robot might first grasp the cloth by an arbitrary point and raise it into the air; it can then detect the lowest point, either while the cloth is held in the air~\cite{cusumano2011bringing,osawa2007unfolding,kita2002model,kita2004deformable,kita2009clothes,mariolis2015pose} or after throwing the cloth on the table~\cite{triantafyllou2011vision}.  
By constraining the cloth to this configuration set, the task of perceiving the cloth or fitting a mesh model~\cite{jimenez2020perception} is greatly simplified. 
However, these funneling actions are usually scripted and  are not generalizable to different cloth shapes or configurations. In contrast, our work aims to enable a robot to interact with cloth from arbitrary configurations and shapes.

Other early works designed vision systems for detecting cloth features that can be used for downstream tasks, such as a Harris Corner Detector~\cite{willimon2011model} or a wrinkle-detector~\cite{sun2013heuristic}. 
More examples of such approaches are described in~\cite{jimenez2020perception}. 
However, these approaches require a task-specific manual design of vision features  and are typically not robust to different variations of the cloth configuration.

\noindent \textbf{Policy Learning for Cloth Manipulation: }
Recently, there have been a number of learning based approaches to cloth folding and smoothing. 
One approach is to learn a policy to achieve a given manipulation task.  Some papers approach this using learning from demonstration.  The demonstrations can be obtained using a heuristic expert~\cite{seita_fabrics_2020} or a scripted sequence of actions based on cloth descriptors~\cite{descriptors_fabrics_2021}.  
Another approach to policy learning is model-free reinforcement learning (RL), which has been applied to cloth manipulation~\cite{matas2018sim,wu2019learning,lee2020learning}. However, policy learning approaches often lack the ability to generalize to novel situations; this is especially problematic for cloth manipulation in which the cloth can be in a wide variety of crumpled configurations.  We compare our method to a state-of-the-art policy learning approach~\cite{wu2019learning} and show greatly improved performance. 

\noindent \textbf{Model-based RL for Cloth Manipulation: }
Model-based RL methods learn a dynamics model and then use it for planning. Model-based reinforcement learning methods have many benefits such as sample efficiency, interpretability, and generalizability to multiple tasks.
Previous works have tried to learn a pixel-based dynamics model that directly predict the future cloth images after an action is applied ~\cite{ebert2018visual,fabric_vsf_2020}. However, learning a visual model for image prediction is difficult and the predicted images are usually blurry, unable to capture the details of the cloth. 
Another approach is to 
represent the cloth with a fixed-size latent vector representation~\cite{yan2020learning} and to plan in that latent space.  However, 
cloth has an intrinsic high dimension state representation; thus,
such compressed representations typically lose the fine-grained details of their environment and are unsuitable for capturing the low-level details of the cloth's shape, such as folds or wrinkles, which can be important for folding or other manipulation tasks. 
Our method also falls into the model-based RL category; unlike previous works, we learn a particle based dynamics model~\cite{li2018learning,sanchez2020learning}, which can better capture the cloth dynamics due to the inductive bias of the particle representation. Additionally, the particle representation is invariant to visual features and enables easier sim-to-real transfer.
\begin{figure*}[h]
    \centering
    \includegraphics[width=\textwidth]{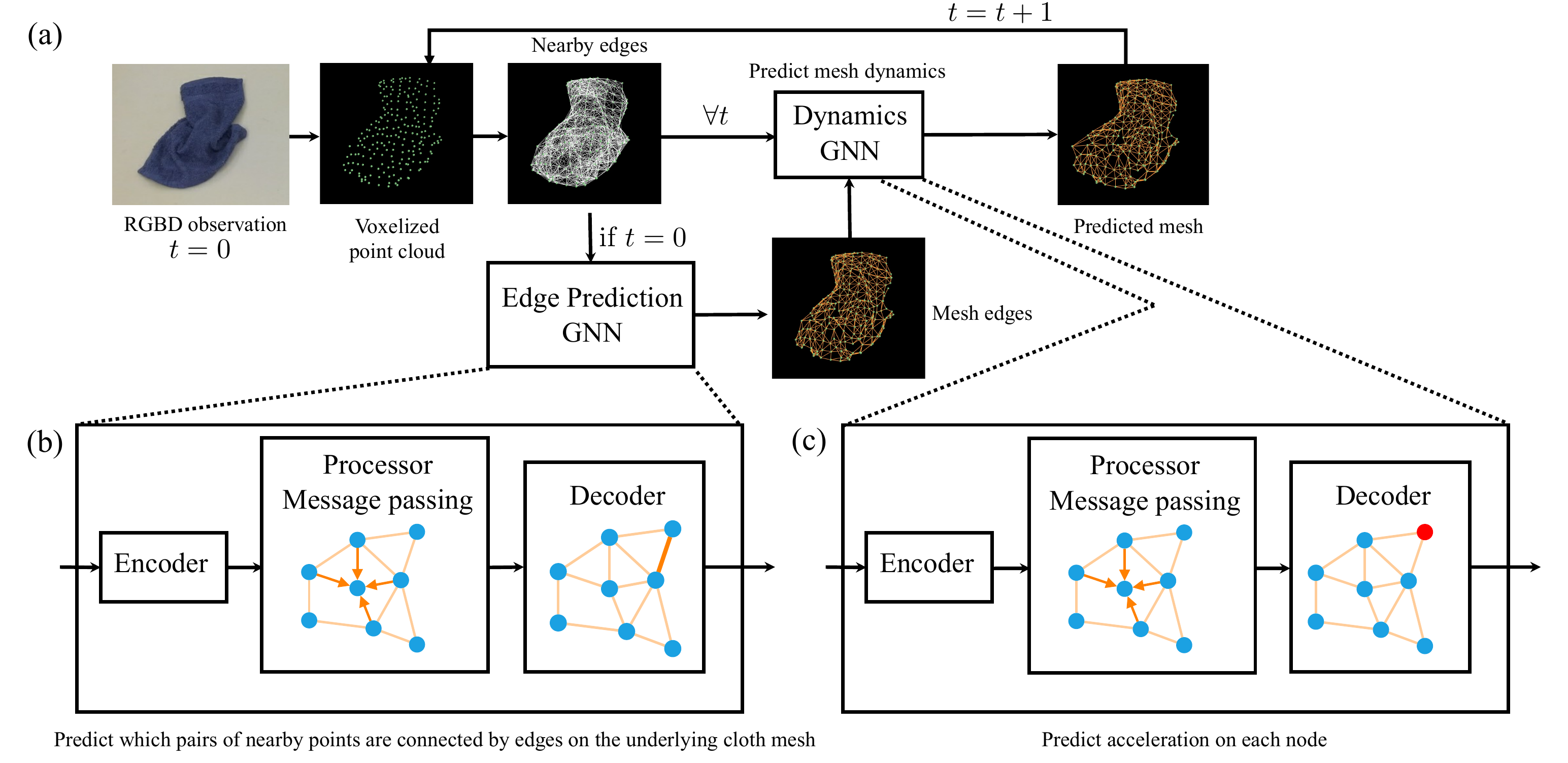}
    \caption{(a) Overview of our visible connectivity dynamics model. It takes in the voxelized point cloud, constructs the mesh and predicts the dynamics for the point cloud. (b) Architecture for the edge prediction GNN which takes in the point cloud connected by the nearby edges and predicts for each nearby edge whether it is a mesh edge. (c) Architecture for the dynamics GNN which takes in the point cloud connected by both the nearby edges and the mesh edges and predict the acceleration of each point in the point cloud.}
    \label{fig:overview}
    \vspace{-5mm}
\end{figure*}
\section{Method}
An overview of our method, \algo~(Visible Connectivity Dynamics), can be found in Figure~\ref{fig:overview}. We represent the cloth using a Visible Connectivity Graph, in which we connect points of a partial point cloud with nearby edges and the inferred mesh edges. Next, we learn a dynamics model over this graph, and finally we use this dynamics model for planning robot actions.



\subsection{Graph Representation of Cloth Dynamics} 
\label{sec:graph-representation}
We represent the state of a cloth with a graph $\langle V, E\rangle$. The nodes $V=\{v_i\}_{i=1...N}$ represent the particles that compose the cloth, where $v_i =(x_i, \dot{x}_i)$
denotes the particle's current position and velocity, respectively.
There are two types of edges $E$ in the graph, representing two types of interactions between the particles:
mesh edges and nearby edges. The mesh edges, $E^M$, represent the connections among the particles on the underlying cloth mesh. The mesh connectivity is determined by the structure of the cloth and does not change throughout time.  Each edge $e_{ij} = (v_i, v_j) \in E^M$ connects nodes $v_i$ to $ v_j$ and models the mesh connection between them. 
The other type of edges are nearby edges, $E^C$, which model the collision dynamics among two particles that are nearby in space. These can be different from the mesh edges due to the folded configuration of the cloth, which can bring two particles close to each other even if they are not connected by a mesh edge. Unlike the mesh edges which stay the same throughout time, these nearby edges are dynamically constructed at each time step based on the following criteria:
\begin{equation}\label{eqn:collision_edge}\small
E^C_t= \big\{e_{ij} \big| ~||x_{i,t}- x_{j,t}||_2 < R \big\},    
\end{equation}
where $R$ is a distance threshold and $x_{i,t}, x_{j,t}$ are the positions of particles $i,j$ at time step t. Throughout the paper, we use the subscript $t$ to denotes the state of a variable at time step $t$ if the variable changes with time.
Additionally, we assume that $E^M \subset E^C$, since a mesh edge connects nodes that are close to each other and hence should also satisfy Eqn.~\ref{eqn:collision_edge}. 

\subsection{Inferring Visible Connectivity from a Partial Point Cloud}
\label{sec:infer_graph_structure}
In the real world, we observe the cloth in the form of a partial point cloud. In this case, we represent the nodes of the graph using the partial point cloud and infer the connectivity among these observed points. We denote the raw point cloud observation as  $P_{raw}=\{x_i\}_{i=1..N_{raw}}$, where $x_i$ is the position of each point and $N_{raw}$ is the number of points. We first pre-process the point cloud by filtering it with a voxel grid filter: we overlay a 3d voxel grid over the observed point cloud and then take the centroid of the points inside each voxel to obtain a voxelized point cloud $P=\{x_i\}_{i=1, ..., N_p}$. This preprocessing step is done both in simulation training and in the real world, which makes our method agnostic to the density of the observed point cloud and more robust during sim2real transfer.

We create a graph node $v_i$ for each point $x_i$ in the voxelized point cloud $P$.  The nearby edges are then constructed by applying the criterion from Eqn.~\ref{eqn:collision_edge}.  However, inferring the mesh edges is less straightforward, since in the real world we cannot directly perceive the underlying cloth mesh connectivity.  To overcome this challenge,
we use a graph neural network~(GNN)~\cite{battaglia2018relational} to infer the mesh edges from the voxelized point cloud. Given the positions of the points in $P$, we first construct a graph $\langle P, E^C \rangle$ with only the nearby edges based on Eqn.~\ref{eqn:collision_edge}. As we assume $E^M \subset E^C$,
we then train a classifier, which is a GNN, to estimate whether each nearby edge $e \in E^C$ is also a mesh edge. We denote this edge GNN as $G_{edge}$. The edge GNN takes as input the graph $\langle P, E^C \rangle$, propagates information along the graph edges in a latent vector space, and finally decodes the latent vectors into a binary prediction for each edge $e\in E^C$ (predicting whether the edge is also a mesh edge). For the edge GNN, we use the network architecture in previous work~\cite{sanchez2020learning} (referred to as GNS). See Appendix A.1 for the detailed architecture. The edge GNN is trained in simulation, where we obtain labels for the mesh edges based on the ground-truth mesh of the simulated cloth. After training, it can then be deployed in the real world to infer the mesh edges from the point cloud. We defer the description of how we obtain the ground-truth mesh labels in Sec.~\ref{sec:train_in_simulator}.


\subsection{Modeling Visible Connectivity Dynamics with a GNN}\label{sec:gnn_dyn}
In order to predict the effect of a robot's action on the cloth, we must model the cloth dynamics. While there exists various physics simulators that support simulation of cloth dynamics~\cite{coumans2019,corl2020softgym,narain2012adaptive}, applying these simulators for a real cloth is still challenging due to two difficulties: first, only a partial point cloud of a crumpled cloth is observed in the real world, usually with many self-occlusions. Second, the estimated mesh edges from Sec.~\ref{sec:infer_graph_structure} may not all be accurate. 
To handle these challenges, we learn a dynamics model based on the voxelized partial point cloud and its inferred visible connectivity (Sec.~\ref{sec:infer_graph_structure}). Formally, given the cloth graph $G_t = \langle V, E \rangle$, a dynamics GNN $G_{dyn}$ predicts the particle accelerations in the next time step, which can then be integrated to update the particle positions and velocities. Here, $V$ refers to the voxelized point cloud, and $E$ refers to inferred visible connectivity that includes both the predicted mesh edges $E^M$ as well as the nearby edges. Our dynamics GNN $G_{dyn}$ uses the similar GNS architecture as the $G_{edge}$. It takes a cloth mesh as input with state information on each node, propagates the information along the graph edges in a latent vector space, and finally decodes the latent vectors into the predicted acceleration on each node. See Appendix A.1 for the detailed architecture of the GNN.



\subsection{Planning with Pick-and-place Actions}
\label{sec:planning}
We plan in a high-level, pick-and-place action space over the VCD model. For each action $a = \{x_{pick}, x_{place}\}$, the gripper grasps the cloth at $x_{pick}$, moves to $x_{place}$, and then drops the cloth. 
As the GNN dynamics model is only trained to predict the changes of the particle states in small time intervals in order to accurately model the interactions among particles, we decompose each high-level action into a sequence of low-level movements, where each low-level movement is a small delta movement of the gripper and can be achieved in a short time. Specifically, we generate a sequence of small delta movements $\Delta x_1, ..., \Delta x_H$ from the high-level action, where $x_{pick} + \sum_{i=1}^H \Delta x_i = x_{place}$. Each delta movement $\Delta x_i$ moves the gripper a small distance along the pick-and-place direction and the motion can be predicted by the dynamics GNN in a single step. When the gripper is grasping the cloth, we denote the picked point as $u$. 
We assume that the picked point is rigidly attached to the gripper; thus, when considering the effect of the $t^{th}$ low-level movement of the robot gripper, we modify the graph by directly setting the picked point $u$'s position $x_{u, t} = x_{pick} + \sum_{i=1}^t \Delta x_i$ and velocity $\dot{x}_{u, t} = \Delta x_i / \Delta t,$ where $\Delta t$ is the time for one low-level movement step. The dynamics GNN will then propagate the effect of the action along the graph when predicting future states. For the initial steps where the historic velocities are not available, we pad them with zeros for input to the dynamics GNN. If no point is picked, e.g., after the gripper releases the picked point, then the dynamics model is rolled out without manually setting any particle state.

Our goal is to smooth a piece of cloth from a crumpled configuration. To compute the reward $r$ based on either the observed or the predicted point cloud, we treat each point in the point cloud as a sphere with radius $R$ and compute the covered area of these spheres when projected onto the ground plane. 
Due to computational limitations, we greedily optimize this reward over the predicted states of the point cloud after a one-step high-level pick-and-place action rather than optimizing over a sequence of pick-and-place actions. Given the current voxelized point cloud of a crumpled cloth $P$, we
first estimate the mesh edges using the edge predictor $E^M = G_{edge}(\langle P, E^C \rangle)$. We keep the mesh edges fixed throughout the rollout of a pick-and-place trajectory since the structure of the cloth is fixed. In theory, it could be helpful to update the mesh edges based on the newly observed point cloud at each low-level step, but this is challenging due to the heavy occlusion from the robot's arm during the execution of a pick-and-place action. After the execution of each pick-and-place action, new particles may be revealed and we update the mesh edges when re-planning the next action. The pseudocode of the planning procedure can be found in the appendix.


\subsection{Training in Simulation}
\label{sec:train_in_simulator}




The simulator we use for training is Nvidia Flex, a particle-based simulator with position-based dynamics~\cite{muller2007position,macklin2014unified}, wrapped in SoftGym~\cite{corl2020softgym}. In Flex, a cloth is modeled as a grid of particles, with spring connections between particles to model the bending and stretching constraints.  

One challenge that we must address is that the points in the observed partial point cloud do not directly correspond to the underlying grid of particles in the cloth simulator.  This presents a challenge for obtaining the ground-truth labels used for training the dynamics GNN and the edge GNN, including the acceleration for each point in the observed point cloud and the mesh edges among them. To address this issue, we perform bipartite graph matching to match each  point in the voxelized point cloud to a simulated particle by minimizing the Euclidean distance between the matched pairs. Details about the matching can be found in the appendix. After we get the mapping from the points to the simulator particles, the ground-truth acceleration of each point is simply assigned to be the acceleration of its mapped particle, which is used for training the dynamics GNN. 
For training the edge GNN, a nearby edge is assumed to be a mesh
edge if the mapped simulation particles of the edge’s both end
points are connected by a spring in the simulator.

\subsection{Graph Imitation Learning for Occlusion Reasoning}
\label{sec:graph_imit}


To better allow our model to reason about occlusions, we introduce graph-based privileged imitation learning, which transfers features from a teacher model trained on the full cloth to a student model trained on the partial cloth observation. 
This idea is related to other recent work which trains a student with partial observations to imitate a teacher with full-state information~\cite{chen2020learning,Lee2020-ff, warrington2020robust}. Specifically, we first train a privileged teacher dynamics model with ground-truth information of the particle state of the full cloth, i.e., it takes as input all particles (including the occluded particles). 
%
Next, we train the student model, which takes a partial point cloud as input.  The student is trained to imitate the features of the corresponding nodes in the teacher~\cite{Lee2020-ff}. 
Graph imitation offers direct supervision on the intermediate features and enables the student model to implicitly reason about the occluded part of the cloth, by imitating the features from the teacher which has full state information. 
We also introduce an auxiliary task of reward (covered area) prediction~\cite{jaderberg2016reinforcement}, which implicitly encodes the cloth shape into the learned features.
Details on our graph-based privileged imitation learning can be found in the appendix. 
\section{Experiments}
\subsection{Experimental Setup}
\label{sec:experiment setup}

\textbf{Simulation Setup}
As mentioned, we use the Nvidia Flex simulator wrapped in SoftGym~\cite{corl2020softgym} for training. The robot gripper is modeled as a spherical picker that can move freely in 3D space and can be activated so the nearest particle will be attached to it. For training, we generate random pick-and-place trajectories on a square cloth. The side length of the cloth varies from 25 to 28 cm.
For evaluation, we consider three different shapes: 1) the same type of square cloth as used in training; 2) Rectangular cloth, with its length and width sampled from $[19, 21] \times [31, 34]$ cm. 3) Two layered T-shirt (the square cloth used for training was single-layered). For each shape, the experiment was run 40 times, each time with a different initial configuration of the fabric.  We report the 25\%, 50\% and 75\% ($Q_{25}, Q_{50}, Q_{75}$) percentiles of the performance. For all our quantitative results, numbers after $\pm$ denotes $\max(|Q_{50} -Q_{25}|, |Q_{75}, - Q_{50}|)$.

Our goal for cloth smoothing is to maximize the covered area of the cloth in the top-down view. We report two performance metrics: Normalized improvement (NI) and normalized coverage (NC). NI computes the increased covered area normalized by the maximum possible improvement ${NI= \frac{s-s_0}{s_{max} - s_0}}$, where $s_0, s, s_{max}$ are the initial, achieved, and maximum possible covered area of the cloth. Similarly, ${NC = \frac{s}{s_{max}}}$ computes the achieved covered area normalized by the maximum possible covered area. We report NI in the main paper and NC in the appendix.

We evaluates two variants of our method: Visible Connectivity Dynamics~(VCD) and VCD with graph imitation learning. We compare with previous state-of-the art methods for cloth smoothing:
VisuoSpatial Foresight (VSF)~\cite{fabric_vsf_2020}, which learns a visual dynamics model using RGBD data; 
Contrastive forward model (CFM)~\cite{yan2020learning}, which learns a latent dynamics model via contrastive learning; 
Maximal Value under Placing (MVP)~\cite{wu2019learning}, which uses model-free reinforcement learning with a specially designed action space. More implementation details can be found in the appendix.



\textbf{Real World Setup}
We use our dynamics model trained in simulation to smooth cloth in the real world with a Franka Emika Panda robot arm and a standard panda gripper, with FrankaInterface library~\cite{zhang2020modular}. We obtain RGBD images from a side view Azure Kinect camera. We use color thresholding for segmenting the cloth and obtain the cloth point cloud. We evaluate on three pieces of cloth: Two square towels made of cotton and silk respectively, and one t-shirt made of cotton. We use our dynamics model trained in simulation without any fine-tuning. More details are in the appendix.

\begin{figure}
    \centering
    \includegraphics[width=0.75\textwidth]{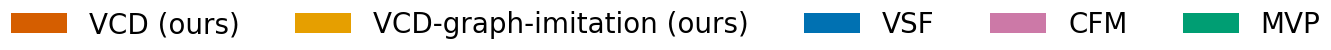}
    \includegraphics[width=.95\textwidth]{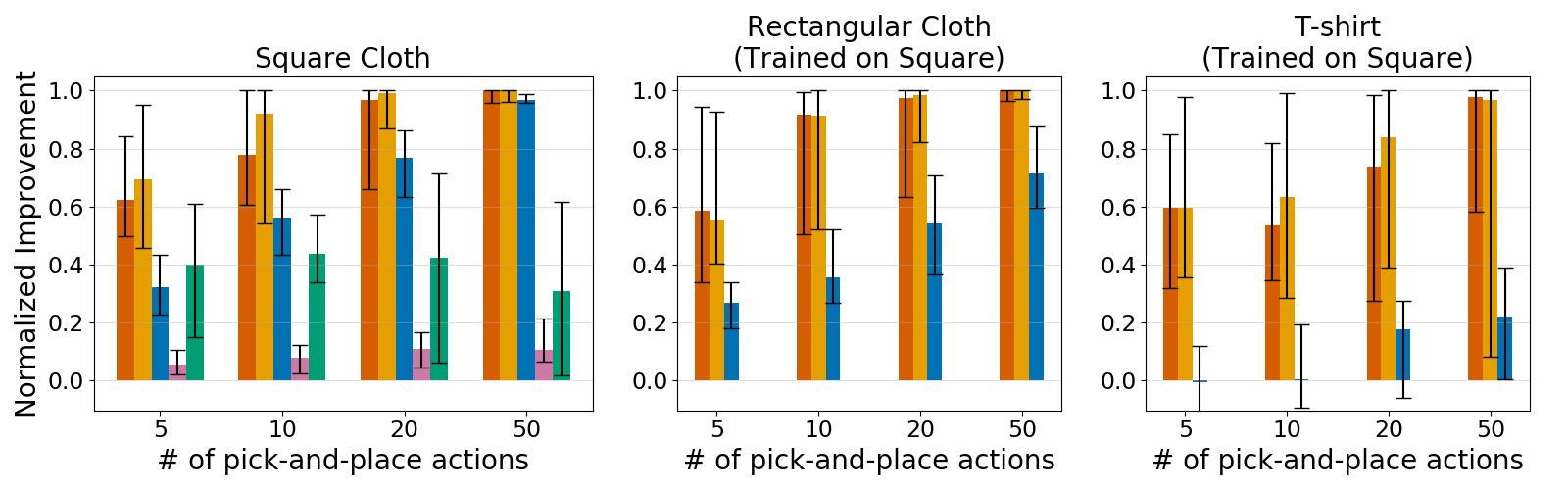}
    \caption{Normalized improvement on square cloth (left), rectangular cloth (middle), and t-shirt (right) for varying number of pick-and-place actions. The height of the bars show the median while the error bars show the 25 and 75 percentile. For detailed numbers, see the appendix.}
    \label{fig:simulation_result}
\end{figure}

\subsection{Simulation Results}
\label{sec:simulation_result}
For each method, we report the NI after different numbers of pick-and-place actions. A smoothing trajectory ends early when NI$>$0.95.  
We note that the edge GNN can achieve a high prediction accuracy of $0.91$ on the validation dataset. See appendix for visualizations of the edge GNN prediction.

We first test all methods on the same type of square cloth used in training. The results are shown in Figure~\ref{fig:simulation_result} (left). 
Under any given number of pick-and-place actions, \algo~greatly outperforms all of the baselines. The graph imitation learning approach described in Section~\ref{sec:graph_imit} further improves the performance.
To test the generalization of these methods to novel cloth shapes that are not seen during training, we further evaluate on a rectangular cloth and a t-shirt. 
For this experiment we only compare VCD to VSF, since VSF achieves the best performance on the square cloth among all the baselines. The results are summarized in Figure~\ref{fig:simulation_result} (middle and right). VCD shows a larger improvement over VSF on the rectangular cloth. T-shirt is more different from the training square cloth and VSF completely fails, while VCD still shows good generalization. The graph imitation learning still leads to marginal improvement and better stability on rectangular since it has a similar shape to the square cloth. However, as the t-shirt has very different shape compared to the square cloth, VCD-graph-imitation does not lead to much improvement and has larger variance on it. 

Since VCD learns a particle-based dynamics model, it incorporates the inductive bias of the cloth structure, which leads to better performance and stronger generalization across cloth shapes, compared to RGB based method like VSF. Please see the appendix for examples of some planned pick-and-place action sequences of our method on all cloth shapes as well as visualizations of the predictions of our model. 





\begin{figure*}[h]
    \centering
    \includegraphics[width=\textwidth]{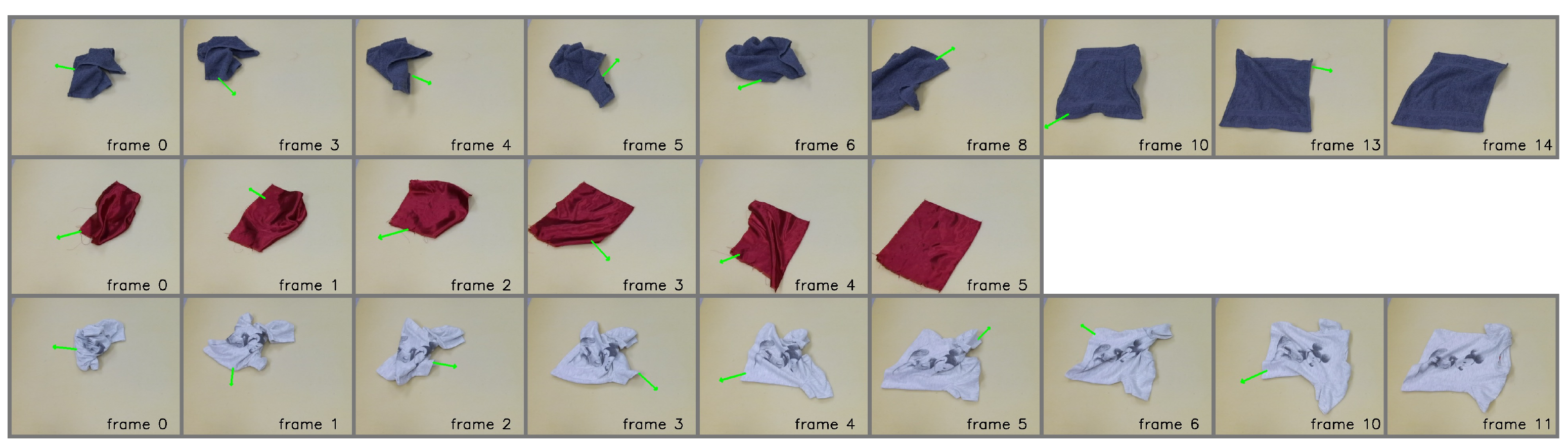}
    \caption{Smoothing cloths of different colors, materials and shapes with our method on a Franka robot: square cotton (top), square silk (middle), cotton t-shirt (bottom).  Each row shows one trajectory. Frame 0 shows the initial configuration of the cloth, and each frame after shows the observation after some number of pick-and-place actions, with the number labeled on the frame. The green arrow shows the 2D projection of the pick-and-place action executed. }
    \label{fig:franka_robot}
\end{figure*}


\begin{table*}[h]
    \centering    \scriptsize
    \begin{tabular}{c|c|c|c|c}

    \toprule
     \diagbox{Material}{\# of pick-and-place\\actions} & 5 & 10 & 20 & Best\\ \hline
    Cotton Square Cloth &  $0.342 \pm 0.265$ & $0.725 \pm 0.445$ & $0.941 \pm 0.360$ & $0.941 \pm 0.153$ \\
Silk Square Cloth & $0.456 \pm 0.197$ & $0.643 \pm 0.391$ & $0.952 \pm 0.229$ & $0.952 \pm 0.095$ \\

Cotton T-Shirt & $0.265 \pm 0.119$ & $0.356 \pm 0.096$ & $0.502 \pm 0.135$ & $0.619 \pm 0.155$ \\

    \bottomrule
    \end{tabular}
    \caption{Normalized improvement of VCD in the real world.}
    \label{tab:robot_performance}
\end{table*}

\subsection{Real-world Results}
We also evaluate our method for smoothing in the real world. We only evaluate VCD (i.e., without graph imitation learning) since it works more stably in simulation. Unfortunately, we were not able to evaluate the baselines in the real world due to the difficulties of transferring their RGB-based policies from simulation.  All of the baselines use RGB data as direct input to the dynamics model or the learned policy, making them sensitive to the camera view and visual features. 
In contrast, our method uses a point cloud as input, which makes it robust to the camera position as well as variation in visual features such as the cloth color or patterns. The point cloud representation allows our method to easily transfer to the real world.

We evaluate 12 trajectories for each cloth. The quantitative results are in Table~\ref{tab:robot_performance} and a visualization of smoothing sequences is shown in Figure~\ref{fig:franka_robot}. Despite the drastic differences of the cotton and silk cloths in visual appearances, shapes, as well as the different dynamics, our model is able to smooth the cotton and silk cloths and generalize well to t-shirt. We also report the performance if our method is able to terminate optimally in hindsight and choose the frame with the highest performance in each trajectory; the result is shown in the last column of Table~\ref{tab:robot_performance}. Videos of  complete trajectories and the model predicted rollouts can be found on our project website. 





\subsection{Ablation Studies}

\begin{wraptable}{r}{7cm}
\centering
\scriptsize
\begin{tabular}{c|c}
\toprule
\multirow{1}{*}{Algorithm} & \multicolumn{1}{c}{Normalized Improvement}  \\ \hline
VCD (Our method) &     $\mathbf{0.778 \pm 0.206}$          \\
Replace dynamics GNN with Flex   & $0.616 \pm 0.143$          \\ 
No edge GNN (dynamic nearby edges)     & $0.531\pm 0.298$          \\ 
No edge GNN (fixed nearby edges)    & $0.599 \pm 0.327$ \\ 
Remove edge GNN at test time      & $0.259\pm 0.118$         \\ 
 \bottomrule
\end{tabular}
    \caption{Normalized improvement of all ablations in simulation after 10 pick-and-place actions.}
    \label{tab:ablation}
\end{wraptable}



We perform the following ablations to study the contribution of each component of our method. 
The first ablation replaces the learned GNN dynamics model with the Flex simulator to test whether a learned dynamics model performs better for our task than the physical simulator. In more detail, after we use the edge GNN to infer the mesh edges on the point cloud, we create a cloth using Flex where a particle is created at each location of the voxelized points and a spring connection is added for each inferred mesh edge. The results is shown in  Table~\ref{tab:ablation}, row 2.
We see that using the Flex simulator instead of the dynamics GNN produces worse performance. The main reason is that the cloth created from the partial point cloud with the inferred mesh edges deviates from the common cloth mesh structure used in Flex; thus, using the Flex simulator under this condition does not create realistic dynamics. Besides, planning with FleX is much slower than planning with the learned GNN dynamics model ($\sim$330s for 1 pick-and-place with FleX and $\sim$40s with VCD).
On the other hand, the dynamics GNN is trained directly on the partial point cloud; 
therefore it can learn to compensate for the partial observability 
when predicting the cloth dynamics. This ablation validates the importance of using a dynamics GNN to learn the dynamics of the partially observable point cloud.



The next set of ablations aims to test whether using an edge GNN to infer the mesh edges as described in Section~\ref{sec:graph-representation} is necessary for learning a good dynamics model. First, we train a dynamics GNN without using the edge GNN, where the edges are constructed solely based on distance by Eqn.~\eqref{eqn:collision_edge}. Since this ablation does not use an edge GNN, it cannot have two different edge types (nearby edges vs mesh edges). Thus at test time, all edges can either be kept fixed throughout the trajectory (similar to the mesh edges in our model), or dynamically reconstructed using Eqn.~\eqref{eqn:collision_edge} at each time step (similar to the nearby edges in our model). The results of these two ablations are shown in Table~\ref{tab:ablation}, rows 3 and 4. As can be seen, the performance is worse without the edge GNN. 

Additionally, we perform another ablation where we train with both nearby edges and mesh edges, but at test time, we do not use an edge GNN to infer the edge type; instead we consider the edges that satisfy the criteria of Eqn.~\eqref{eqn:collision_edge} in the first time step as the mesh edges. The result of this ablation is shown in Table~\ref{tab:ablation}, row 5. The performance is again much worse. 
All these ablations validate the importance of using an edge GNN to infer the mesh edges.



\section{Conclusion}
In this paper, we propose the visible connectivity dynamics~(VCD) model, that infers a visibility connectivity graph from the partial point cloud and learns a particle-based dynamics model over the graph for planning to perform cloth smoothing. \algo~has the advantage of posing strong inductive bias that fits the underlying cloth physics, being invariant to visual features, and being interpretable.  We show that \algo~greatly outperforms previous state-of-the-art methods for cloth smoothing, and achieves zero-shot sim-to-real transfer on a Franka arm for smoothing various types of cloth. 

Our work demonstrates the importance of the choice of state representation for efficient and generalizable manipulation, as well as the benefits of a graph-based representation. While there may not be a universal representation suitable for all objects, we believe that a graph representation can be an important alternative to raw images or latent vectors, especially for deformable objects such as cloth. In the future, we hope VCD can also be applied to different types of object manipulation tasks, such as manipulation of cables, bags, and food.




\clearpage
\acknowledgments{We would like to thank Wen-Hsuan Chu for his initial Pytorch implementation of the GNS model. We would like to thank members of the RPAD lab, Gokul Swamy and Erica Weng for their feedback on the early draft of the paper. This material is based upon work supported by the National Science Foundation under Grant No. IIS-2046491 and LG Electronics.}


\bibliography{supplement}  

\appendix
\clearpage
\newpage
\renewcommand{\figurename}{Supplementary Figure}
\renewcommand{\tablename}{Supplementary Table}
\addcontentsline{toc}{section}{Appendix} 
\part{Appendix} 
\parttoc 

\section{VCD Implementation}
\subsection{GNN Architecture}
\label{sec:GNN Architecture}
As mentioned in the main paper, we take the network architecture in previous work~\cite{sanchez2020learning} (referred to as GNS) for our dynamics GNN $G_{dyn}$ and the edge GNN $G_{edge}$. Both GNN consists of three parts: encoder, processor and decoder. Since the dynamics and edge GNNs have very similar architectures, we first describe the architecture of the dynamics GNN, and then describe how the edge GNN architecture differs from that.

\textbf{Input: }The input to the dynamics GNN is a graph, where the nodes are the points in the voxelized point cloud of the cloth, and the edges consist of the collision edges (built using Eq. (1)) and mesh edges (inferred by a trained edge GNN). The node feature for a point $v_i$ consists of the concatenation of its past $m$ velocities, a one-hot encoding of the point type (picked or unpicked 
- see details about picking in Section 3.4 in the main paper and Section~\ref{sup:planning} in the appendix), and the distance to the table plane. 
For edge $e_{jk}$ that connects nodes $v_j$ and $v_k$, its edge feature consists of the distance vector
$(x_j - x_k$), its norm $||x_j - x_k||$, 
 a one-hot encoding of the edge type (mesh edge or collision edge), and the current displacement from the rest position  $||x_j - x_k|| - r_{jk}$, where $r_{jk}$ is the distance between $x_j$ and $x_k$ at the rest positions. The displacement from the rest positions are set to zero for collision edges which do not have rest positions.

We now describe how the robot action is incorporated into the input graph of the dynamics GNN as follows.
As mentioned in Section 3.4 of the main paper, when we want to use the dynamics GNN to predict the effect of a pick-and-place robot action $a = \{a_{pick}, a_{place}\}$ on the current cloth, we first decompose the high-level action into a sequence of low-level movements, where each low-level movement is a small delta movement of the gripper and can be achieved in a short time. Specifically, we generate a sequence of small delta movements $\Delta x_1, ..., \Delta x_H$ from the high-level action, where $x_{pick} + \sum_{i=1}^H \Delta x_i = x_{place}$. Each delta movement $\Delta x_i$ moves the gripper a small distance along the pick-and-place direction and the motion can be predicted by the dynamics GNN in a single step. We then incorporate the small delta movement into the input graph as follows. When the gripper is grasping the cloth, we denote the picked point as $u$. We assume that the picked point is rigidly attached to the gripper; thus, when considering the effect of the $t^{th}$ low-level movement of the robot gripper, we modify the input graph by directly setting the picked point $u$'s position $x_{u, t} = x_{pick} + \sum_{i=1}^t \Delta x_i$ and velocity $\dot{x}_{u, t} = \Delta x_i / \Delta t,$ where $\Delta t$ is the time for one low-level movement step. The dynamics GNN will then propagate the effect of the robot action along the graph when predicting future states. 

\textbf{Encoder:} The encoder consists of two separate multi-layer perceptrons~(MLP), denoted as $\phi_p, \phi_e$, that map the node and edge feature, respectively, into latent embedding. Specifically, the node encoder $\phi_p$ maps the node feature for node $v_i$ into the node embedding $h_i$, and the edge encoder $\phi_e$ maps the edge feature for edge $e_{jk}$ into the edge embedding $g_{jk}$.

\textbf{Processor:}
The processor consists of $L$ stacked Graph Network~(GN) blocks~\cite{battaglia2018relational} that update the node and edge embedding, with residual connections between blocks. We use $L=10$ in both edge GNN $G_{edge}$ and dynamics GNN $G_{dyn}$. The $l^{th}$ GN block contains an edge update MLP $f^l_e$ and a node update MLP $f^l_p$ that take as input the edge and node embedding $g^l$ and $h^l$ respectively and outputs updated embedding $g^{l+1}$ and $h^{l+1}$ (we denote $g^0$ and $h^0$ as the edge and node embedding output by the encoder). It also contains a global update MLP $f_c^{l}$ that takes as input a global vector embedding $c^l$, and outputs the updated global embedding $c^{l+1}$. The initial global embedding $c^0$ is set to be 0. For each GN block, first the edge update MLP updates the edge embedding; it takes as input the current edge embedding $g^l_{jk}$, the node embedding $h^l_j, h^l_k$ for the nodes that it connects, as well as the global embedding $c^l$: 
$
g_{jk}^{l+1} = f^l_e(h^l_j, h^l_k, g^l_{jk}, c^l) + g^l_{jk}, ~~\forall e_{jk} \in E. 
$
The node update MLP then updates the node embedding; its input consists of the current node embedding $h_i^l$, the sum of the updated edge embedding for the edges that connect to the node, and the global embedding $c^l$: 
$
        h_i^{l+1} = f^l_p(h_i^l, \sum_j g_{ji}^{l+1}, c^l) + h^l_i, ~~\forall i = 1, ..., N_p.
$
Note the edge and node updates both have residual connections between consecutive blocks.
Finally, the global update MLP takes as input the current global embedding $c^l$, the mean of the updated node and edge embedding, and updates the global embedding as: $c^{l+1} = f_c^l(c^l, \frac{1}{|V|} \sum_{i=1}^{|V|} h_i^{l+1}, \frac{1}{|E|} \sum_{e_{jk}} g_{jk}^{l+1})$.

\textbf{Decoder:}
The decoder is an MLP $\psi$ that takes as input the final node embedding $h^L_i$ output by the processor for each point $v_i$; the decoder outputs the acceleration for each point:
$
    \ddot{x}_i = \psi(h^L_i).
$
The acceleration can then be integrated using the Euler method to update the node position $x_i$.
We train the graph GNN $G_{dyn}$ using the L2 loss between the predicted point acceleration $\ddot{x}_i$ and the ground-truth acceleration obtained by the simulator; see Sec.~\ref{sup:VCDtrain} for details.

\textbf{Edge GNN:} The edge GNN $G_{edge}$ has nearly the same architecture as the dynamics GNN, with the following differences: first, the input graph to the edge GNN encoder consists of only the voxelized point cloud and the collision edges $\langle P, E^C \rangle$; the edge GNN aims to infer which collision edges are also mesh edges. The node feature is 0 for all nodes. The edge feature for edge $e_{jk}$ consists of the distance vector $(x_j - x_k)$ and its norm $||x_j - x_k||$ (without the edge type, since this must be inferred by the edge GNN). The processor is exactly the same as that in the dynamics GNN. The decoder is an MLP that takes as input  the final edge embedding output by the processor and outputs the probability of the collision edge being a mesh edge. We use a binary classification loss on the prediction of the mesh edge for training.

\textbf{Hyperparameters}
In simulator, we set the radius of particles to be 0.00625, an
All MLPs that we use has three hidden layers with 128 neurons each and use ReLU as the activation function. The detailed parameters of the GNN architecture, as well as the simulator parameters, can be found in \tablename~\ref{tab:hyper_params}.

\subsection{VCD Training Details}
\label{sup:VCDtrain}

\hspace{4mm} \textbf{Details about training in simulation:}
We trian the dynamics GNN with one-step prediction loss: suppose that we sample a transition $(V_t, a_t, V_{t+1})$, where $a_t$ is a low-level action.  Then we assign the velocity at timestep $t$ that is input to the network to be the ground-truth velocity obtained from the simulator (after matching the points to their corresponding simulator particles). This strategy enables us to sample arbitrary timesteps for training rather than needing to always simulate the dynamics from the first timestep.

For training the edge GNN, we need to obtain the ground-truth of which collision edges are also mesh edges.  During simulation training,
a collision edge is assumed to be a mesh edge if the mapped simulation particles of the edge's both end points are connected by a spring in the simulator.

We train our dynamics GNN with the ground-truth mesh edges, and directly use it with the mesh edges predicted by the edge GNN at test time. We find this to work well without fine-tuning the dynamics GNN on mesh edges predicted by the edge GNN, due to the high prediction accuracy(91\%) of the edge GNN.

\textbf{Details about bipartite graph matching:}
\label{sup:VCDtrain:bipartite}
As mentioned in the main paper, we need bi-partite graph matching to find a mapping from the voxelized point cloud to the simulation particles, in order to obtain the state and connectivity of the voxelized point cloud for training the dynamics and edge GNN.
Given $N$ points in the voxelized point cloud $p_i, i=1\dots N$ and $M$ simulated particles of the cloth in simulation $x_j, j=1\dots M$, the goal of the bipartite graph matching here is to match each point in the point cloud to a simulated particles. The simulated cloth mesh is downsampled by three times to improve computation efficiency, e.g., a cloth composed of $40 \times 40$ particles is downsampled to be of size $13 \times 13$. The bi-partite matching is only performed on the downsampled particles. We build the bipartite graph by connecting an edge from each $p_i$ to $x_j$, with the cost of the edge being the distance between the two points. In our experiments, we always have $M>N$ since we use a large grid size for the voxelization. 

\textbf{Training data: } We collect 2000 trajectories, each consisting of 1 pick-and-place action. The pick point is randomly chosen among the locally highest points on the cloth; this is only done to generate the training data for the dynamics model, not for planning (we do this for training the VSF and CFM baselines as well; the MVP baseline uses the behavioral policy to generate its training data). The unnormalized direction vector $p = (\Delta x, \Delta y, \Delta z)$ for the pick-and-place action is uniformly sampled as follows: $\Delta x, \Delta z \in [-0.5, 0.5]$, $\Delta y \in [0, 0.5]$. The direction vector is then normalized and the move distance is sampled uniformly from $[0.15, 0.4]$. The high-level pick-and-place action is decomposed into $100$ low-level steps: the pick-and-place is executed in the 60 low-level actions, and then we wait 40 steps for the cloth to stablize. We train our dynamics model in terms of low-level actions.

We choose the voxel size (0.0216) to be three times of the particle radius (0.00625) to keep it consistent with the downsampled mesh. The neighbor radius, which determines the construction of collision edges, is set to be roughly two times of the voxel size, so as to ensure that particles in adjacent voxels are connected.


\textbf{Training parameters: }
\label{sup:VCDtrain:procedure}
We use Adam~\cite{kingma2014adam} with an initial learning rate of 0.0001 and reduce it by a factor of 0.8 if the training plateaus. We train with a batch size of 16.
The training of the dynamics GNN takes roughly 4 days to converge on a RTX 2080 Ti. The training of the edge GNN usually converges in 1 or 2 days. Detailed training parameters can be found in \tablename~\ref{tab:hyper_params}.

\begin{table*}[h!]\centering
\begin{tabular}{@{}lp{40mm}}
\toprule
Model parameter & Value\\
\midrule
\textit{Encoder(same for both node encoder and edge encoder)} & \\
\hspace{5mm}number of hidden layers & 3 \\
\hspace{5mm}size of hidden layers & 128 \\

\textit{Processor} & \\
\hspace{5mm}number of message passing steps & 10 \\
\hspace{5mm}number of hidden layers in each edge/node update MLP  & 3 \\
\hspace{5mm}size of hidden layers & 128 \\

\textit{Decoder} & \\
\hspace{5mm}number of hidden layers & 3 \\
\hspace{5mm}size of hidden layers & 128 \\

\toprule
Training parameters & Value\\
\midrule
\hspace{5mm}learning rate & 0.0001\\
\hspace{5mm}batch size & 16 \\
\hspace{5mm}training epoch & 120\\
\hspace{5mm}optimizer & Adam\\
\hspace{5mm}beta1 & 0.9\\
\hspace{5mm}beta2 & 0.999\\
\hspace{5mm}weight decay & 0\\



\toprule
Others & Value\\
\midrule
\hspace{5mm}dt  & 0.05 second\\
\hspace{5mm}particle radius & 0.00625 m\\
\hspace{5mm}downsample scale & 3\\
\hspace{5mm}voxel size & 0.0216 m\\
\hspace{5mm}neighbor radius $R$ & 0.045 m\\
\bottomrule
\end{tabular}
\caption{Summary of all hyper-parameters.}
\label{tab:hyper_params}
\end{table*}

\newpage
\subsection{Graph-imitation Training Details}
\begin{figure*}[h]
    \centering
    \includegraphics[width=\textwidth]{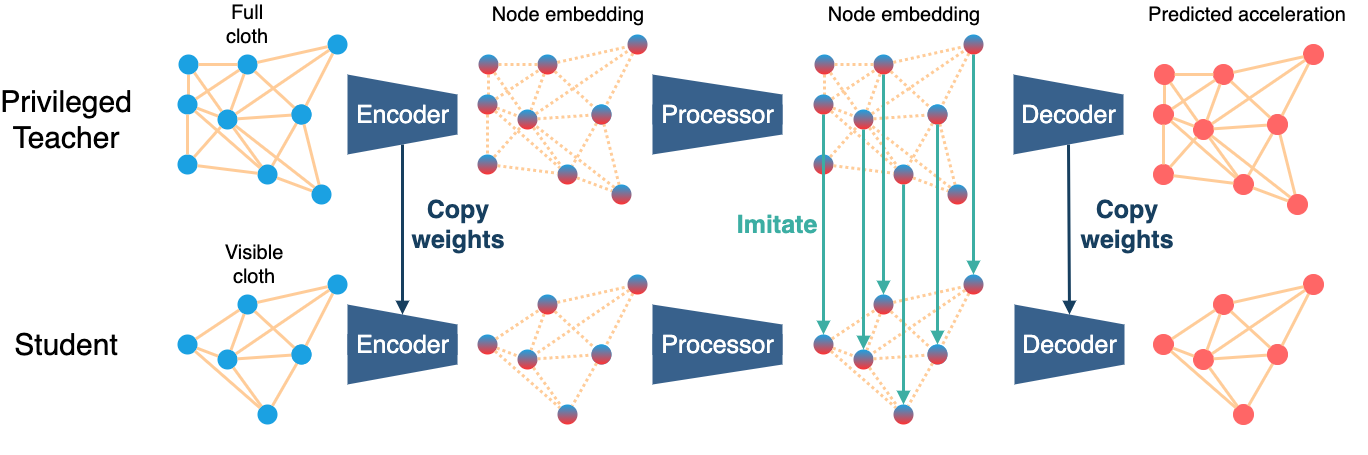}
    \caption{A graphical illustration of privileged graph imitation learning. The privileged teacher has the same model architecture as student, but takes full cloth as input. Following \cite{Lee2020-ff}, we initialize the encoder and decoder of student model by weights of pretrained teacher. Then we freeze the teacher and transfer the privileged information by matching the node embedding and global embedding of two models. The target nodes to imitate are obtained by bi-partite matching as described in \ref{sup:VCDtrain:bipartite}.}
    \label{fig:graph_imit}
\end{figure*}
Although VCD performs decently under partial observability, we found dynamics model trained on full mesh model usually converges faster and obtains better asymptotic performance. This is well expected since incomplete information caused by self-occlusion results in ambiguity of state estimation. 

Therefore, we introduce graph-imitation learning to inject prior knowledge of the full cloth into the dynamics model. The prior encodes structure of the full cloth and incentivizes the model to reason about occlusion implicitly.

\subsubsection{Privileged Graph-imitation Learning}
The main spirit of privileged graph-imitation learning is to train a student model which takes partial point cloud as input, to imitate a privileged agent which has access to privileged information. We hope the student to learn a recover function that recovers true states from partial information. A visual illustration is shown in \figurename~\ref{fig:graph_imit}.

To do so, we first train a privileged agent with all simulated particles(including the occluded ones) and ground-truth mesh edges. The privileged teacher model shares identical architecture as the student model, but with complete information. We train the teacher model with acceleration loss and the auxiliary reward prediction loss.

Graph-based imitation learning is not straightforward because the graphs of two models have very different structures. Typically, the graph of teacher model will have more vertices since it can observe occluded particles while the student only observes the voxelized partial point cloud. To tackle with this challenge, we conduct bipartite matching to match student nodes with teacher nodes as described in \ref{sup:VCDtrain:bipartite}.

Once we have the node correspondence, we retrieve the intermediate node features of both teacher and student model, $h_T^L$ and $h_S^L$, and force the node feature of student $h_S^L$ to be similar to $h_T^L$. The final output is still supervised by groundtruth acceleration. We copy the weights of encoder and decoder from teacher model to initialize student since we find it accelerate training. The teacher is frozen during imitation learning. By imitating the intermediate node features, we provide high capacity training signal to the student to recover groundtruth acceleration by proper message passing. To successfully imitate the teacher, the student have to conduct occlusion reasoning to some extent, and take the effects of occluded particles and erroneous mesh edges into account. In addition to node features, the student model also mimic the global embedding of teacher model to make more accurate reward prediction. We use mean square error for imitation learning.

\subsubsection{Auxiliary Reward Prediction}
Following \cite{jaderberg2016reinforcement}, we additionally train our dynamics model to predict reward in order to regularize the model. The groundtruth reward, which is the coverage of cloth after one time step, is calculated by approximation as described in \ref{sup:planning}. The coverage is calculated over all particles, thus it provides information from a global view to the model. The reward model is a three-layer MLP which takes global embedding $c^L$ as input. We use mean square error to train the model.

It should be noted that at test time, we still use the heuristic reward function, which models particles as spheres and calculate an approximate coverage on the partial point cloud. Although the learned reward model predicts a global reward, which theoretically will take into the newly revealed occluded regions into account,  we found it perform slightly worse than the heuristic reward function.

\subsection{VCD Planning Details}
\label{sup:planning}

We summarize the planning procedure of VCD in Algorithm~\ref{algo:planning}.  We sample $K$ high-level pick-and-place actions. For each sampled high-level action, we roll out our dynamics model using that action for $H$ low-level steps and obtain the sequence of predicted point positions. 

\begin{algorithm}
\SetAlgoLined
\DontPrintSemicolon 
 \SetKwInOut{Input}{input}
 \SetKwInOut{Output}{output}
 \Input{Voxelized partial point cloud $P$, Edge GNN $G_{edge}$, Dynamics GNN $G_{dyn}$, number of sampled actions $K$} 
 \Output{pick-and-place action $a = \{x_{pick}$, $x_{place}\}$}
 Build collision edges $E^C_0$ with $P$;
 Infer mesh edges $E^M \gets G_{edge}(P, E^C_0)$\;
 \For{$i \gets 1$ \textbf{to} $K$} 
 {

    Sample a pick-and-place action $x_{pick}$, $x_{place}$ \;  
    Compute low-level actions $\Delta x_1, ..., \Delta x_H$\; 
    Get picked point $v_u$ from $x_{pick}$\;
    Pad historic velocities with 0:
    $\mathbf{x}_0 \gets P, \dot{\mathbf{x}}_{-m...0} \gets \mathbf{0}$\;
    \For {$t \gets 0$ \textbf{to} $H$}
    {
        Build collision edges $E^C_t$ with $\mathbf{x}_t$\;
        Move  picked point according to gripper movement by : \; 
        $x_{u, t}\gets x_{u, t} + \Delta x_t$, \hspace{0.5pt} $\dot{x}_{u, t} \gets \Delta x_t / \Delta t $\;
        Predict accelerations using $G_{dyn}$:
        $\ddot{\mathbf{x}}_{t} \gets G_{dyn} (\mathbf{x}_t, \dot{\mathbf{x}}_{{t-m}...t}, u, E^M, E^C_t)$\;
        Update point cloud predicted positions \& velocities: \;
        $\dot{\mathbf{x}}_{t+1} = \dot{\mathbf{x}}_{t} + \ddot{\mathbf{x}}_{t}  \Delta t$, \hspace{0.5pt}
        $\mathbf{x}_{t+1} = \mathbf{x}_{t} + {\dot{\mathbf{x}}}_{t+1}  \Delta t$\;
         Readjust picked point according to gripper movement  by \;
         $x_{u, t}\gets x_{u,t} + \Delta x_t$, \hspace{0.5pt} $\dot{x}_{u, t} \gets \Delta x_t / \Delta t $\;
    }
    Compute reward $r$ based on final point cloud predicted position $\mathbf{x}_H$\;
 }
 \Return{pick and place action with maximal reward}\;
 \caption{Planning with pick-and-place actions}
 \label{algo:planning}
\end{algorithm}

\textbf{Action sampling during planning in simulation}
As described in the main text, we sample 500 pick-and-place actions, where the pick point is first uniformly sampled from a bounding box of the cloth and then projected to be on the cloth mask. For generating the bounding box, we first obtain the cloth mask from the simulator. We then obtain the minimal and maximal pixel coordinates $u, v$ value of the cloth mask. The bounding box is the rectangle with corners $(min(u) - padding, min(v) - padding)$ and $(max(u) + padding, max(v) + padding)$, where padding is set to be 30 pixels for the $360 \times 360$ image size we use. 
We use rejection sampling to make sure the place point is within the image to keep the action within the depth camera view.
    The unnormalized direction vector $p = (\Delta x, \Delta y, \Delta z)$ ($y$ is the up axis) of the pick-and-place is uniformly sampled as follows: $\Delta x, \Delta z \in [-0.5, 0.5]$, and $\Delta y \in [0, 0.5]$. The vector is normalized and then the distance is separately sampled from $[0.05, 0.2]$ meters. We decompose the pick-and-place action into 10 low-level actions and wait for another 6 steps for the cloth to stabilize. 

\textbf{Action sampling during planning in the real world}
The robot action space is pick-and-place with a top down pinch grasp. For each action, we sample 100 pick-and-place actions to be evaluated by our model. Each action sample is generated as follows: We first sample a pick-point location corresponding to the segmented cloth, denoted as $(p_x, p_y)$.  We then generate a random direction $\theta \in [0, 2\pi]$ and distance $l \in [0.02, 0.1]$ meters. 
Then the place point will be $(p_x + l \cos \theta), p_y + l\sin \theta$.  We only accept an action if both the pick and the place points are within the work space of the robot. We additionally filter out actions whose place points are overlapping with the cloth. This heuristic saves computation time without sacrificing performance.

\textbf{Reward computation in planning: }
\label{planning:reward}
As described in the main text, to compute the reward function $r$ for planning, we treat each node in the graph as a sphere with radius $R$ and compute the covered area of these spheres when projected onto the ground plane. To prevent the planner from exploiting the model inaccuracies, we do the following: if the model predicts that there are still points above a certain height threshold after executing the pick-and-place action and waiting the cloth to stabilize, then the model must be predicting inaccurately and we set the reward of such actions to 0. The threshold we use is computed as $15 \times 0.00625$ meters, where $0.00625$ is the radius of the cloth particle used in the simulation. 

\section{Baselines Implementation}

For all the baselines, we try our best to adjust the SoftGym cloth environment to match the cloth environment used in the original papers. For VSF, we place the camera to be top-down and zoomed in so that the cloth  covers the entire image when fully flattened. We also changed the color of the cloth to be bluish as in the original paper. We collect 7115 trajectories, each consisting of 15 pick-and-place actions for training the VSF model (same as in the VSF paper). For CFM, we also use a top-down camera and change the color of the cloth to be the same on both sides, following the suggestion of the authors (personal communication). We collect 8000 trajectories each consisting of 50 pick-and-place actions for training the contrastive forward model (same as in the CFM paper). For MVP, we collect 5000 trajectories each with 50 pick-and-place actions and report the performance of the best performing model during training. 
We trained each of the baselines for at least as many  pick-and-place actions as they were trained in their original papers.
For training our method, \algo, we collect 2000 trajectories, each consisting of 1 pick-and-place action decomposed into 20 low-level actions for training.  Note that this is fewer pick-and-place actions than any of the baselines used for training. We now describe each compared baseline in more details below:

\subsection{VisuoSpatial Forsight (VSF)}
We use the official code of VSF provided by the authors\footnote{https://github.com/ryanhoque/fabric-vsf}.

\begin{figure*}
    \centering
    \begin{tabular}{cccc}
    \includegraphics[width=0.2\textwidth]{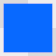} &
    
    \includegraphics[width=0.2\textwidth]{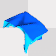} &
    
    \includegraphics[width=0.2\textwidth]{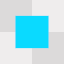} &
    
    \includegraphics[width=0.2\textwidth]{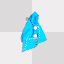} \\
    (a) VSF goal image. & (b) VSF observation. & (c) CFM goal image. & (d) CFM observation. 
    
    \end{tabular}
    \caption{Images of cloth configurations used by the baseline methods.}
    \label{fig:baseline-imgs}
\end{figure*}

\textbf{Image: } Following the original paper, we use images of size $56 \times 56$. we place the camera to be top-down and zoomed in so that the cloth  covers the entire image when fully flattened. An example goal image of the smoothed cloth for VSF is shown in \figurename~\ref{fig:baseline-imgs}.  

\textbf{Training data \& Procedure: } For training the VSF model, we collect 7115 trajectories for training (same as in the VSF paper), each consisting of 15 pick-and-place actions. Following the VSF paper, the pick-and-place action
first moves the cloth up to a fixed height, which is set to be 0.02 m in our case, and then moves horizontally. The horizontal movement vector is sampled from $[-0.07, 0.07] \times [-0.07, 0.07]$ m. This range is smaller than what is used for VCD, as we follow the original paper to set the maximal move distance roughly  half of the cloth/workspace size. 
We use rejection sampling to ensure the after the movement, the place point is within the camera view. Similar to VCD, the pick point is uniformly sampled among the locally highest points on cloth (only during training). It takes 2 weeks for VSF to converge on this dataset. 

\textbf{Action sampling during planning:}
Similarly to VCD, the pick point is sampled uniformly from a bounding box around the cloth and then projected to the cloth mask. The padding for the bounding box here we use is $6$. Other than the pick point, other elements of the pick-and-place action is sampled following the exact same distribution as in the training data collection. 

\subsection{Contrastive Forward Model (CFM)}
We use the official code of CFM provided by the authors\footnote{https://github.com/wilson1yan/contrastive-forward-model}.

\textbf{Image: } Following the original paper, we use images of size $64 \times 64$. We also place the camera to be top-down and adjust the camera height so the cloth contains a similar portion of the image as in the original paper. Following the suggestions from the authors (personal communication), we also set the color of the cloth to be the same on both sides. See \figurename~\ref{fig:baseline-imgs} for an example of the images we use.

\textbf{Training data: }  For training, we collect 8000 trajectories each consisting of 50 pick-and-place actions, which is the same as in the original paper. Similar to VCD and VSF, the pick point is sampled among the locally highest points on the cloth (only during training). The movement vector is sampled from $[-0.04, 0.04] \times [0, 0.04] \times [-0.04, 0.04]$ m, where the y-axis is the negative gravity direction. We use pick-and-place actions with such small distances following the original paper. We also use rejection sampling to ensure the place point is within the camera view.

\textbf{Action sampling during planning: }
Similar to VCD, the pick point is sampled uniformly from a bounding box around the cloth and then projected to the cloth mask. The padding size here we use for the bounding box is $5$. Other than the pick point, other elements of the pick-and-place action are sampled following the exact same distribution as in training data collection.

\subsection{Maximal Value under Placing (MVP)}
We use the official code of MVP provided by the authors\footnote{https://github.com/wilson1yan/rlpyt}.

\textbf{Image: } Following the original paper, we use images of size $64 \times 64$. We also place the camera to be top-down.

\textbf{Training data: }  For training, we collect 8000 trajectories each consisting of 50 pick-and-place actions, which is the same as the original paper. However, the Q function starts to diverge after 5000 trajectories and the performance starts to drop. Thus we report the best policy performance when it has been trained for 5000 trajectories. This corresponds to around 15000 training iterations.

\textbf{Action space: } The action space for the MVP policy is in 5 dimension: $(u, v, \Delta x, \Delta y, \Delta z)$, where $u, v$ is the image coordinate of the pick point  and is sampled for the segmented cloth pixel. We use the depth information to back project the pick point to 3d space. $(\Delta x, \Delta y, \Delta z)$ is the displacement of the place location relative to the pick point and is clipped to be within $0.5$. Additionally, the height $\Delta y$ is clipped to be non-negative.
    

\section{Experimental Setup}
 
\subsection{Simulation Setup}

\begin{figure}
    \centering
    \includegraphics[width=0.3\textwidth]{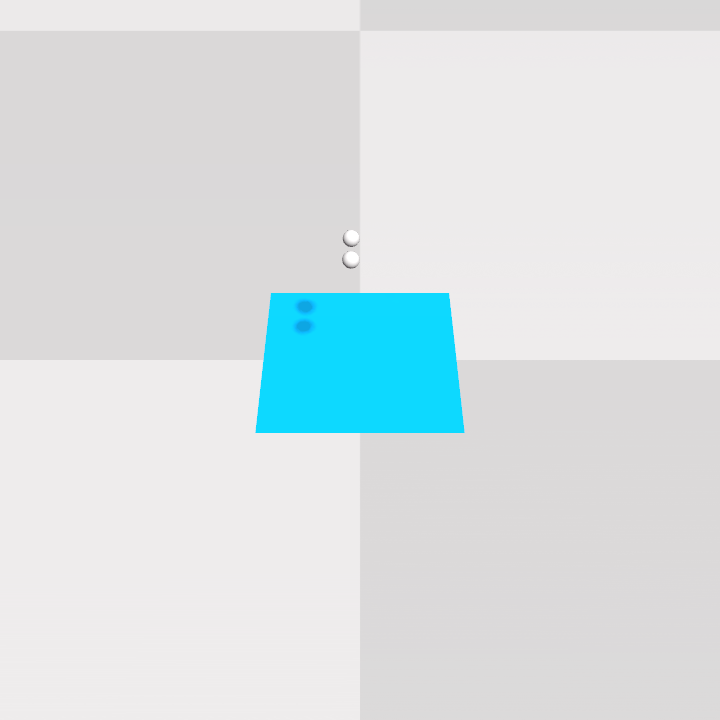}
    \includegraphics[width=0.3\textwidth]{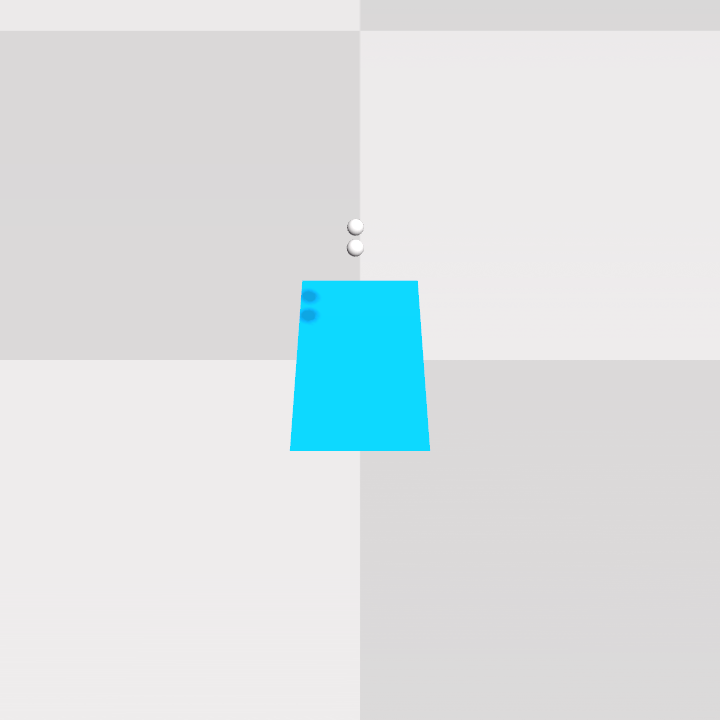}
    \includegraphics[width=0.3\textwidth]{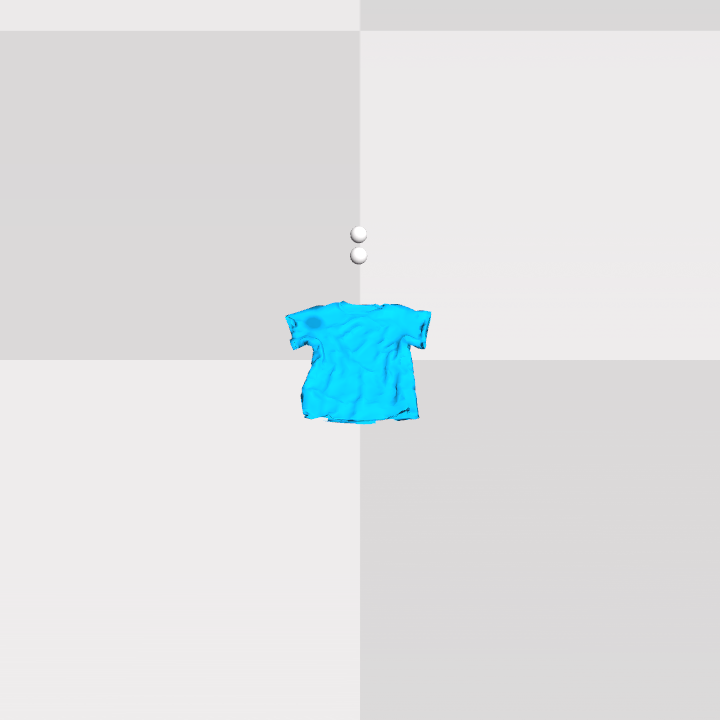}
    \caption{Images of the square cloth, rectangular cloth, and t-shirt used in simulation.}
    \label{fig:simulation_cloth}
\end{figure}

We use the Nvidia Flex simulator, wrapped in SoftGym~\cite{corl2020softgym}, for training. In SoftGym, the robot gripper is modeled using a spherical picker that can move freely in 3D space and can be activated so the nearest particle will be attached to it. For the simulation experiments, we use a nearly square cloth, composed of a variable number of particles sampled from $[40, 45] \times [40, 45]$; this corresponds to a cloth of size in the range of $[25, 28] \times [25, 28]$ cm. Detailed cloth parameters such as stiffness are listed in the appendix. For all methods, we randomly generate 20 initial cloth configurations for training. The initial configurations are generated by picking the cloth up and then dropping it on the table in simulation. For evaluation, we consider three different geometries: 1) the same type of square cloth as used in training; 2) Rectangular cloth. The length and width of the rectangular cloth is sampled from $[19, 21] \times [31, 34]$ cm. 3) T-shirt.  Images of these three shapes of cloth in simulation are shown in \figurename~\ref{fig:simulation_cloth}.

We set the stiffness of the stretch, bend, and shear spring connections to  $0.8, 1, 0.9$, respectively.

\subsection{Real-world Setup}
\textbf{Real World Setup}
Our real robot experiments use a Franka Emika Panda robot arm with a standard panda gripper. We obtain RGBD from a side view Azure Kinect camera and crop the RGBD image into the size of [345, 425], which corresponds to a workspace of 0.4 x 0.5 meters. To obtain the cloth point cloud, we first use color thresholding to remove the table background and obtain the cloth segmentation mask and then back project each cloth pixel to 3d space using the depth information. We evaluate on three pieces of cloth: Two squared towels made of silk and cotton respectively and one shirt made of cotton. We use the covered area as described in Sec.~\ref{sup:planning} as our reward function. 

For each cloth, we evaluate 12 trajectories each with a maximum of 20 pick-and-place actions. For each trajectory, the robot stops if the normalized performance is higher than 0.95 or if the predicted rewards of all the sampled actions are smaller than the current reward.  For each trajectory, we reset the cloth configuration using the following protocol: Each time, the arm picks a random point on the cloth, lifts it up to 0.4 meters above the table and drop it at a fixed point on the table. This procedure is done three times in the beginning of each trajectory.

\section{Additional Experimental Results}

\subsection{Simulation Experiments}
\subsubsection{Normalized Improvement and Normalized Coverage in Simulation}
NI and NC of our simulation experiments are reported in \tablename~\ref{tab:simulation_performance} and \tablename~\ref{tab:simulation_performance_nc}. With different metrics, our method consistently outperforms all baselines.

\begin{table*}[h]
    \centering
    \scriptsize
    \begin{tabular}{c|c|c|c|c|c}
    \toprule
     & \diagbox{Method}{\# of pick-and-\\place actions} & 5 & 10 & 20 & 50 \\ 
     \bottomrule
    \multirow{5}{*}{Square} & \algo ~(Ours)    & $0.624\pm0.217$ &  $0.778\pm0.222$ &  $0.968\pm0.307$ & $1.000\pm0.043$  \\
    & \algo-graph-imitation (Ours)    & $\mathbf{0.692\pm0.258}$ &  $\mathbf{ 0.919\pm0.377 }$ &  $\mathbf{ 0.990\pm0.122}$ & $\mathbf{1.000\pm0.039}$  \\
    & VSF~\cite{fabric_vsf_2020}     & $0.321 \pm 0.112$ &  $0.561 \pm 0.127$ & $0.767 \pm0.134$ & $0.968 \pm 0.021$ \\
    & CFM~\cite{yan2020learning} & $0.053 \pm 0.051$ &  $0.077 \pm 0.053$ & $0.109 \pm 0.066$ & $0.105 \pm 0.106$   \\
    & MVP~\cite{wu2019learning} & $ 0.399\pm 0.210$ & $ 0.435\pm 0.137$ & $ 0.421\pm 0.361$ & $0.307 \pm 0.310$   \\
    \bottomrule
    \multirow{3}{*}{Rectangular} & \algo ~(Ours)    & $\mathbf{0.585\pm0.359}$ &  $\mathbf{0.918\pm0.413}$ &  $ 0.973\pm0.341$ & $0.979\pm0.399$  \\
    & \algo-graph-imitation (Ours)    & $0.556\pm0.372$ &  $0.912\pm0.393$ &  $\mathbf{0.985\pm0.164}$ & $\mathbf{1.000\pm0.028}$  \\
    & VSF~\cite{fabric_vsf_2020}     & $0.268 \pm 0.090$ &  $0.356 \pm 0.163$ & $0.542 \pm 0.177$ & $0.715 \pm 0.162$ \\
    \bottomrule
    \multirow{3}{*}{T-shirt} & \algo ~(Ours)    & $\mathbf{0.595\pm0.279}$ &  $ 0.533\pm0.285$ &  $ 0.738\pm0.465$ & $\mathbf{0.979\pm0.399}$  \\
    & \algo-graph-imitation (Ours)    & $0.595\pm0.385$ & $\mathbf{0.633\pm0.357}$  &  $\mathbf{0.838\pm0.450}$ & $ 0.969\pm0.8860$  \\
    & VSF~\cite{fabric_vsf_2020}     & $-0.009 \pm 0.125$ &  $0.004 \pm 0.188$ & $0.176 \pm 0.237$ & $0.219 \pm 0.218$ \\
    \bottomrule
    \end{tabular}
    \caption{Normalized Improvement (NI) of all methods in simulation, for varying numbers of allowed pick and place actions.}
    \label{tab:simulation_performance}
\end{table*}

\begin{table*}[h]
    \centering
    \scriptsize
    \begin{tabular}{c|c|c|c|c|c}
    \toprule
     & \diagbox{Method}{\# of pick-and-\\place actions} & 5 & 10 & 20 & 50\\ 
     \bottomrule
    \multirow{5}{*}{Square} & \algo ~(Ours)    & $0.776\pm0.132$ &  $0.872\pm0.128$ &  $0.985\pm0.1873$ & $1.000\pm0.023$  \\
    & \algo-graph-imitation (Ours)    & $\mathbf{0.837\pm0.150}$ &  $\mathbf{0.966\pm0.236}$ &  $\mathbf{0.994\pm0.076}$ & $\mathbf{1.000\pm0.021}$  \\
    & VSF~\cite{fabric_vsf_2020}     & $0.629 \pm 0.053$ &  $0.762 \pm 0.093$ & $0.878 \pm 0.090$ & $0.984 \pm 0.010$  \\
    & CFM~\cite{yan2020learning} & $0.445 \pm 0.052$ &  $0.466 \pm 0.044$ & $0.494 \pm 0.031$ & $0.538 \pm 0.044$  \\
    & MVP~\cite{wu2019learning} & $ 0.667\pm 0.121$ & $ 0.667\pm 0.124$ & $ 0.661\pm 0.194$ & $0.609 \pm 0.179$  \\
    \bottomrule
    \multirow{3}{*}{Rectangular} & \algo ~(Ours)    & $\mathbf{0.785\pm0.182}$ &  $\mathbf{0.957\pm0.233}$ &  $ 0.985\pm0.183$ & $0.998\pm0.017$  \\
    & \algo-graph-imitation (Ours)    & $0.768\pm0.191$ &  $0.949\pm0.215$ &  $\mathbf{0.992\pm0.080}$ & $\mathbf{1.000\pm0.015}$  \\
    & VSF~\cite{fabric_vsf_2020}     & $0.622 \pm 0.078$ &  $0.664 \pm 0.078$ & $0.765 \pm 0.119$ & $0.860 \pm 0.072$  \\
    \bottomrule
    \multirow{3}{*}{T-shirt} & \algo ~(Ours)    & $0.837\pm0.107$ &  $ 0.828\pm0.096$ &  $0.897\pm0.150$ & $\mathbf{0.991\pm0.189}$  \\
    & \algo-graph-imitation (Ours)    & $\mathbf{0.867\pm0.143}$ &  $\mathbf{0.901\pm0.179}$ &  $\mathbf{0.960\pm0.218}$ & $ 0.991\pm0.331$  \\
    & VSF~\cite{fabric_vsf_2020}     & $0.636 \pm 0.086$ &  $0.653 \pm 0.090$ & $0.676 \pm 0.079$ & $0.698 \pm 0.075$  \\
    \bottomrule
    \end{tabular}
    \caption{Normalized coverage (NC) of all methods in simulation on the regular cloth, for varying numbers of allowed pick and place actions.}
    \label{tab:simulation_performance_nc}
\end{table*}

\subsubsection{Visualization of Edge GNN}
\begin{figure}[h]
    \centering
    \includegraphics[width=.4\textwidth ]{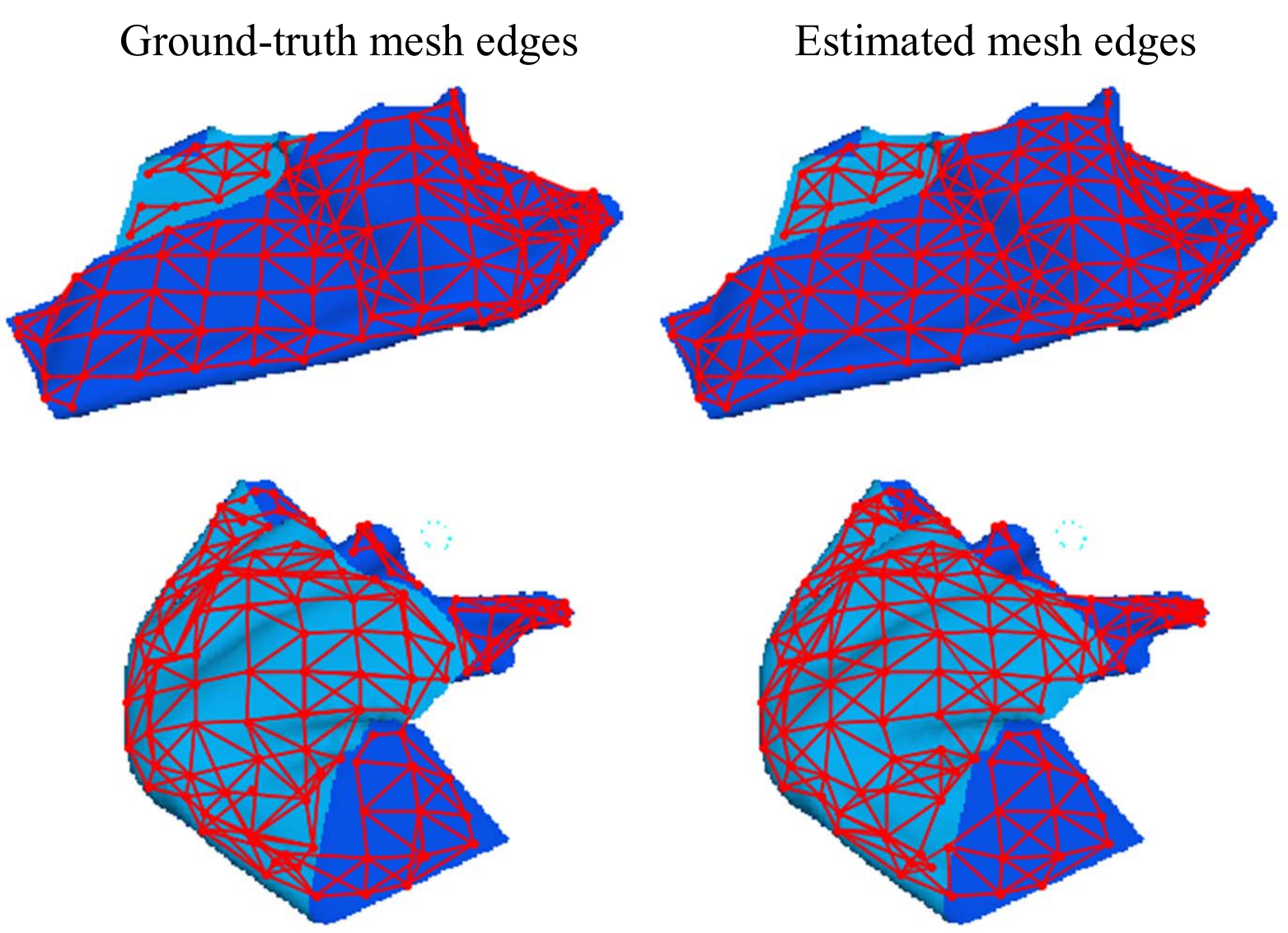}
    \caption{The edge prediction result of our edge GNN. Red lines visualize the ground-truth (left) or inferred (right) mesh connections.}
    \label{fig:edge-prediction}
\end{figure}

We compare predictions of our edge prediction model with the ground-truth edges used for training the edge model in \figurename~\ref{fig:edge-prediction}. As shown, the edge GNN prediction reasonably matches the ground-truth, and thus well captures the cloth structure; it can also correctly disconnect the top layer from the bottom layer when the cloth is folded, e.g., the top left part of the first example and the bottom right part of the second example. Note our method uses only the point cloud as input and the color in this figure is only used for visualization. 
The edge GNN is trained on the same dataset as the dynamics GNN (described in Sec.~\ref{sup:VCDtrain:procedure}), and on the validation set, it achieves a prediction accuracy of $0.91$.


\subsubsection{Visualizations of Planned Actions in Simulation}

%

\begin{figure*}[h]
    \centering
        \includegraphics[width=\textwidth]{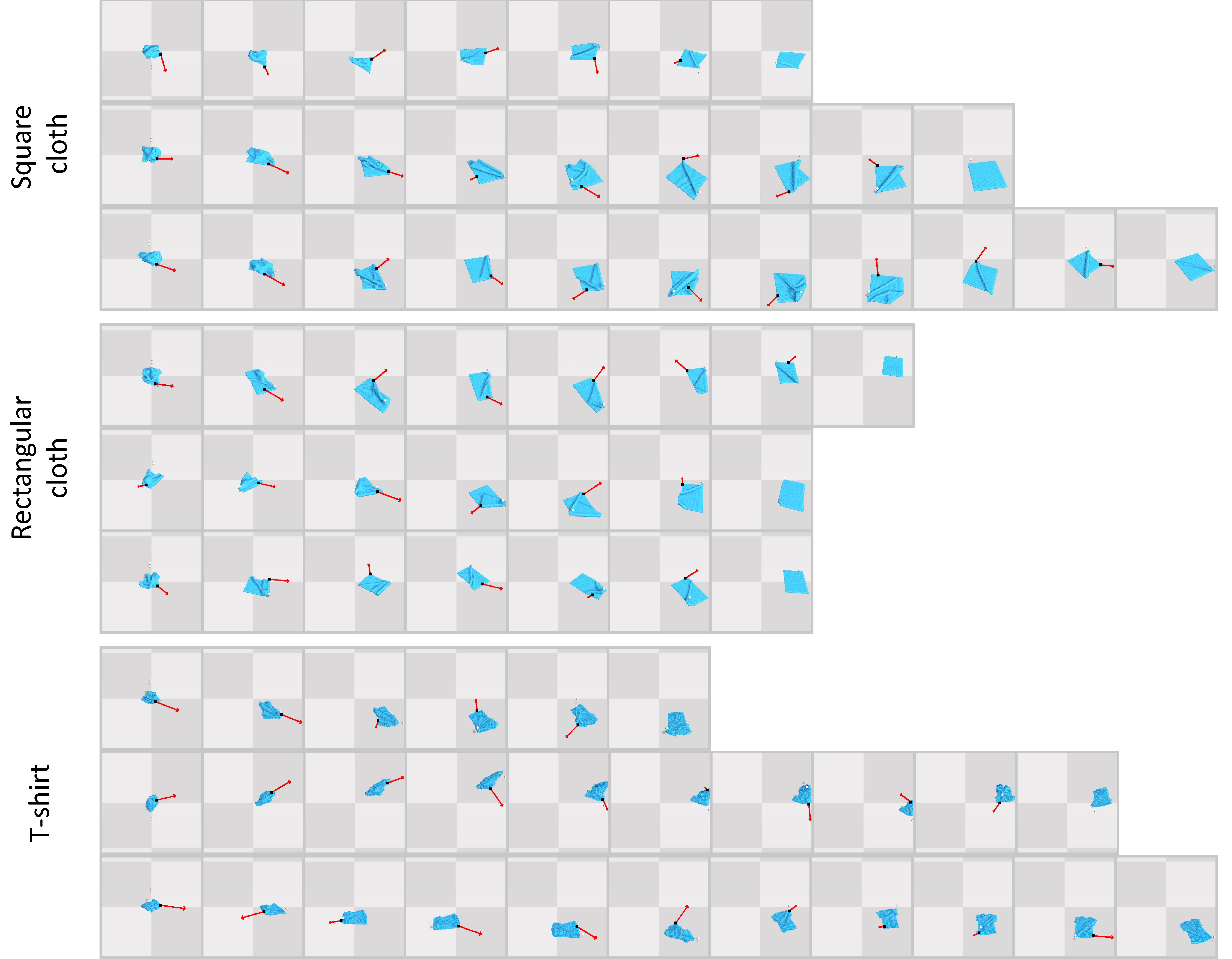}
    \caption{Three example planned pick-and-place action sequences for square cloth, rectangular cloth, and t-shirt. All trajectories shown achieve a normalized improvement above 0.98.}
    \label{fig:planned-pick-and-place}
\end{figure*}

\figurename~\ref{fig:planned-pick-and-place} shows three planned pick-and-place action sequences of \algo~in simulation. As shown, VCD successfully plans actions that gradually smooths the cloth. We observe note that VCD favours picking edge / corner points and pulling outwards, which is an effective smoothing strategy, demonstrating the effectiveness of VCD for planning.

\begin{figure*}[h]
    \centering
    \includegraphics[width=0.8\textwidth]{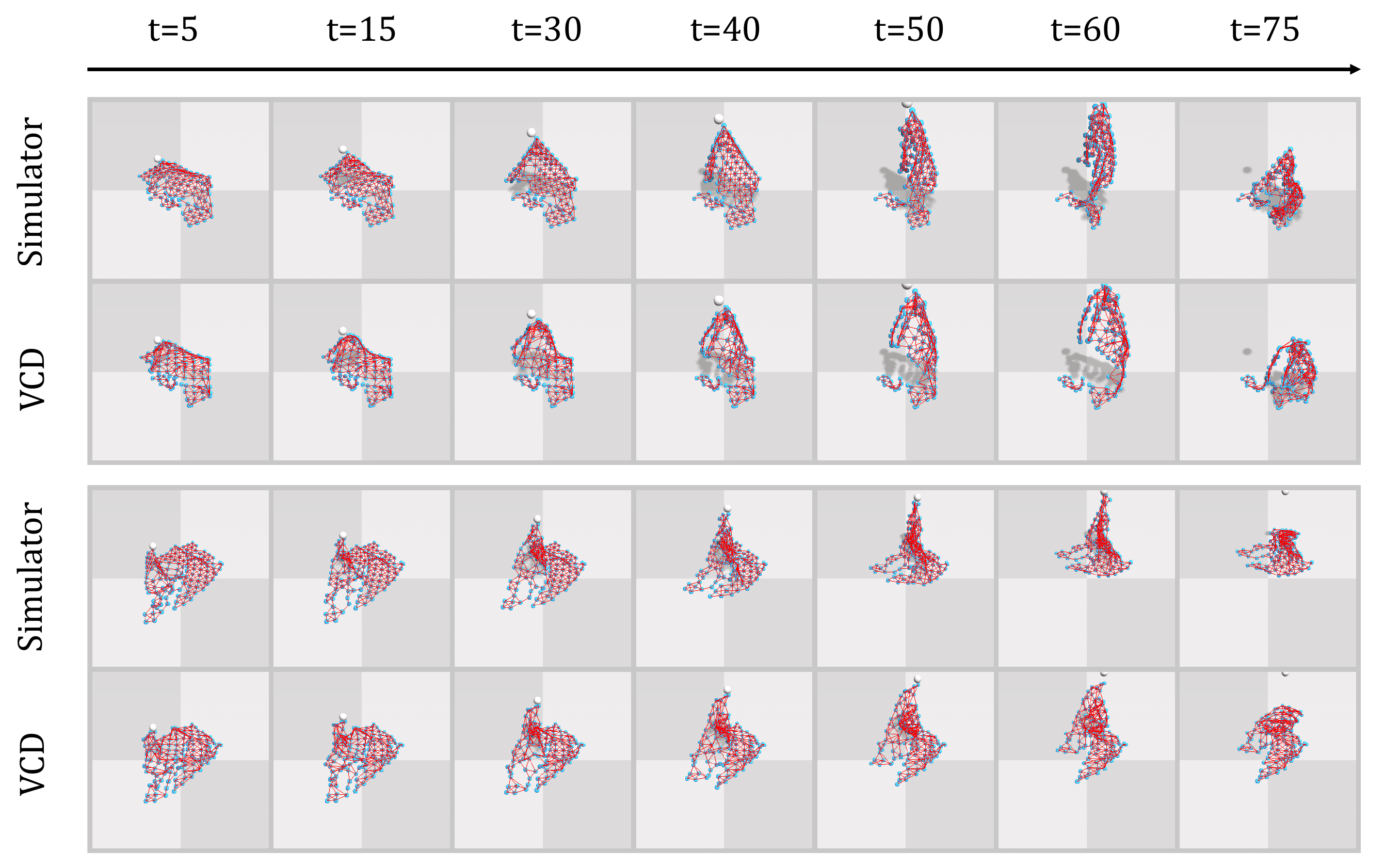}
    
    \caption{Two open-loop predictions of \algo~ on square cloth. Blue points are observable particles/point cloud points and red lines are mesh edges. For each prediction, the top row is the ground-truth observable particles connected by the ground-truth mesh edges in simulator. The bottom row is the predicted point clouds by \algo, in which the mesh edges are inferred by the edge prediction GNN.}
    \label{fig:open-loop-prediction-square}
\end{figure*}

\begin{figure*}[h]
    \centering
    \includegraphics[width=0.8\textwidth]{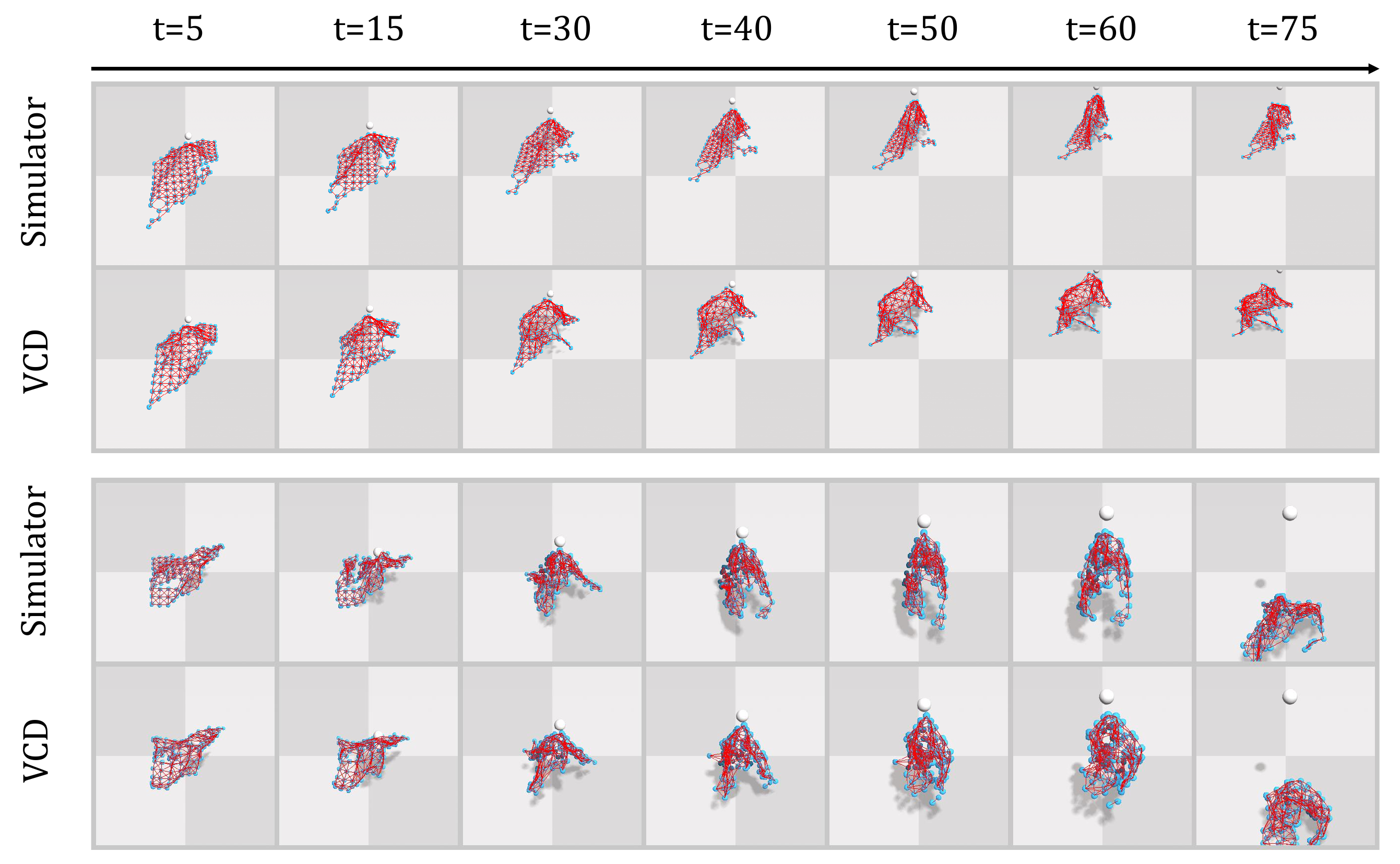}
    
    \caption{Two open-loop predictions of \algo~ on rectangular cloth. Note VCD is only trained on square cloth. Blue points are particles/point cloud points and red lines are mesh edges. For each prediction, the top row is the ground-truth observable particles connected by the ground-truth mesh edges in simulator. The bottom row is the predicted point clouds by \algo, in which the mesh edges are inferred by the edge prediction GNN.}
    \label{fig:open-loop-prediction-rectangular}
\end{figure*}

\begin{figure*}[h]
    \centering
    \includegraphics[width=0.8\textwidth]{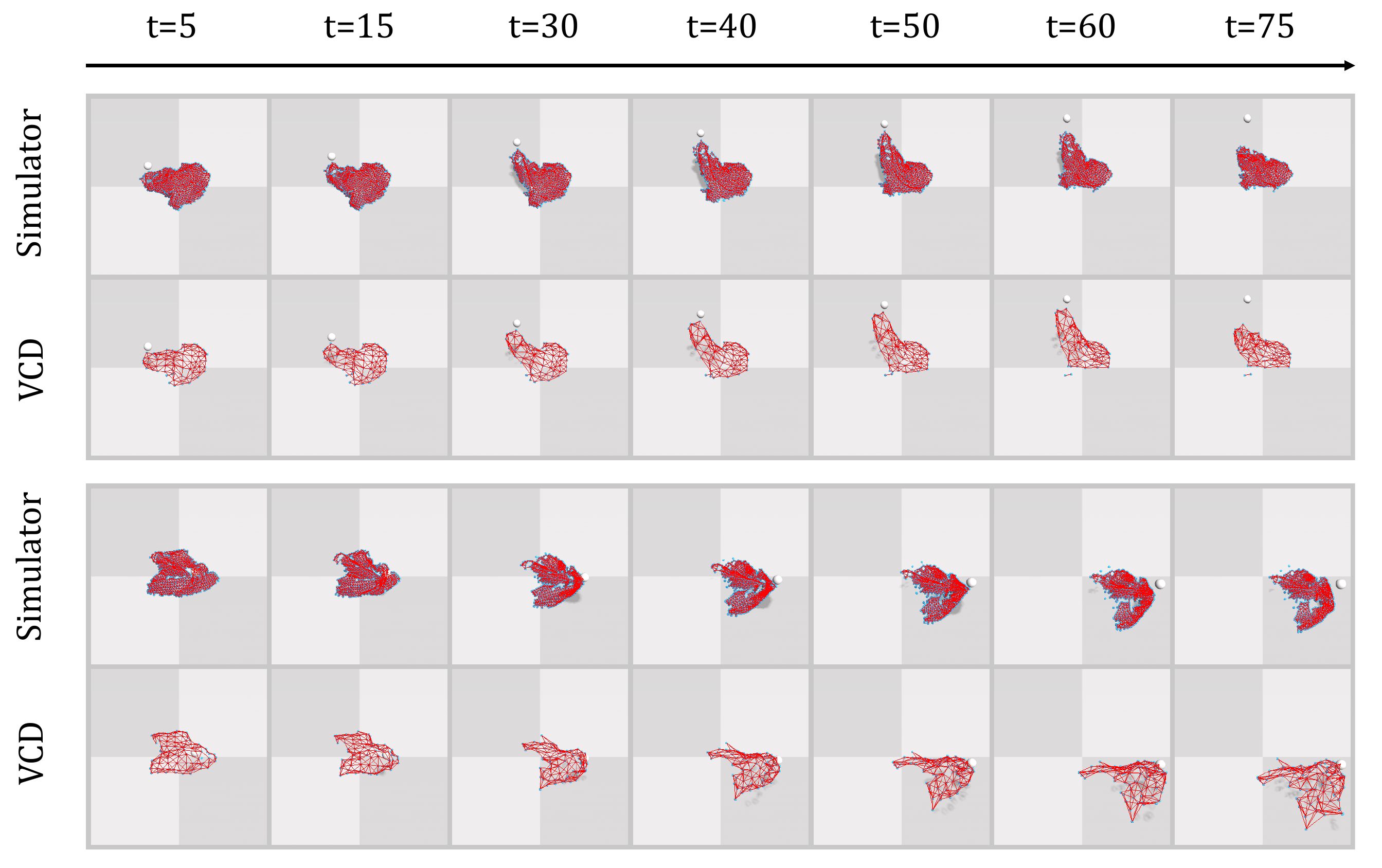}
    
    \caption{Two open-loop predictions of \algo~ on t-shirt. Blue points are particles/point cloud points and red lines are mesh edges. Note VCD is only trained on square cloth. For each prediction, the top row is the ground-truth observable particles connected by the ground-truth mesh edges in simulator. The bottom row is the predicted point clouds by \algo, in which the mesh edges are inferred by the edge prediction GNN.}
    \label{fig:open-loop-prediction-tshirt}
\end{figure*}



\subsubsection{Visualizations of Open-loop Predictions in Simulation}

In order to understand better what our model is learning, we visualize the prediction of our model compared to the simulator output in \figurename~\ref{fig:open-loop-prediction-square}, \ref{fig:open-loop-prediction-rectangular}, \ref{fig:open-loop-prediction-tshirt}. Given a pick-and-place action decomposed into 75 low-level actions, the model is given the $5^{th}$ point cloud in the trajectory with the past 4 historical velocities, and the dynamics model is used to generate the future predictions. As shown, even if the prediction horizon is as long as 70 steps, \algo~is able to give relatively accurate predictions on all cloth shapes, indicating the effectiveness of incorporating the inductive bias of the cloth structure into the dynamics model.




\subsection{Robot Experiments}
\subsubsection{Running Time}
In average, it takes 12.7 seconds for VCD to plan each pick-and-place action (100 samples) on 4 RTX 2080Ti and 10.2 seconds for Franka to execute the action. With additional communication overhead, our current system takes around 40 seconds for computing and executing each pick-and-place action.

\subsubsection{Normalized Coverage (NC) of Robot Experiments}
For the robot experiments, the main text reports the normalized improvement~(NI). NC are reported here in \tablename~\ref{tab:robot_performance2}.  

\begin{table*}[h]
    \centering    \scriptsize
    \begin{tabular}{c|c|c|c|c}

    \toprule
     \diagbox{Material}{\# of pick-and-place\\actions} & 5 & 10 & 20 & Best\\ \hline
    Cotton Square Cloth &  $0.690 \pm 0.166$ & $0.884 \pm 0.293$ & $0.959 \pm 0.193$ & $0.959 \pm 0.080$ \\
Silk Square Cloth & $0.744 \pm 0.180$ & $0.876 \pm 0.314$ & $0.964 \pm 0.075$ & $0.964 \pm 0.054$ \\

Cotton T-Shirt & $0.548 \pm 0.114$ & $0.601 \pm 0.093$ & $0.688 \pm 0.068$ & $0.773 \pm 0.141$ \\

    \bottomrule
    \end{tabular}
    \caption{Normalized coverage~(NC) of VCD in the real world.}
    \label{tab:robot_performance2}
\end{table*}

\subsubsection{Visualization of Sampled Actions in The Real World}
\begin{figure}[h]
    \centering
    \includegraphics[width=0.8\textwidth]{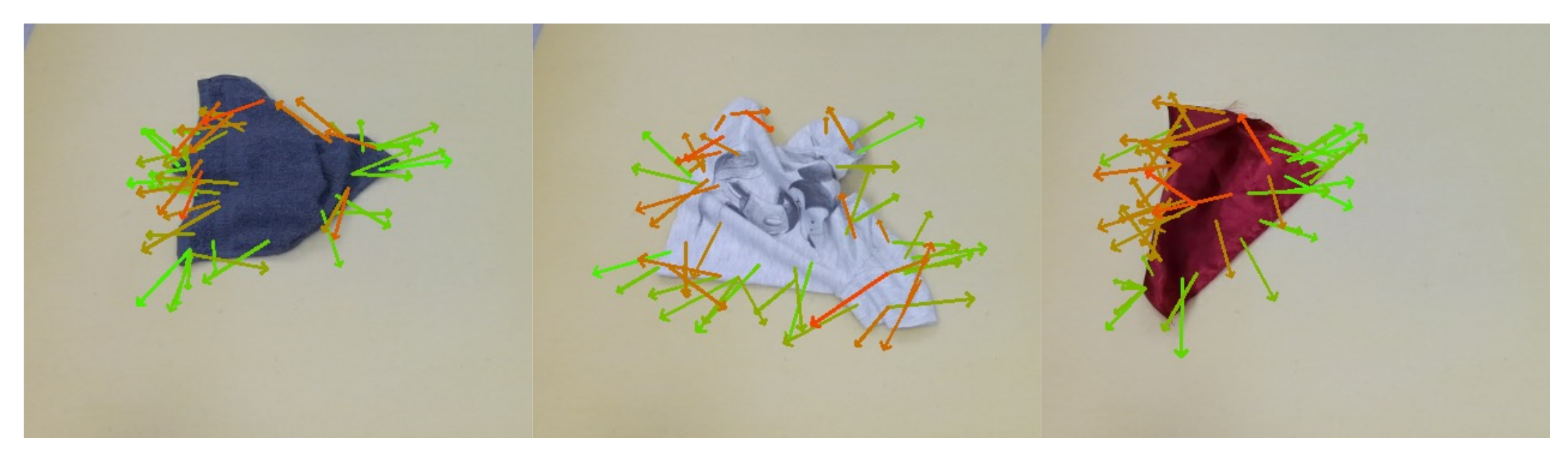}
    \caption{Examples of 50 sampled actions used for planning. Each arrow goes from the 2D projection of the pick location to that of the place location. The actions with the higher predicted reward are shown in greener color and the actions with the lower predicted reward are shown in redder colors. }
    \label{fig:robot_sampled_actions}
\end{figure}

We show in \figurename~\ref{fig:robot_sampled_actions} VCD's predicted score for each of the sampled action during smoothing of the cloth. Interestingly, though there is no explicit optimization for this, \algo~favours picking corner or edge points and pulling outwards, which is a very natural and effective strategy for smoothing. This demonstrates the effectiveness of \algo~for planning.

\clearpage
\section{Planning with VCD for Cloth Folding}
\begin{figure}
    \centering
    \begin{tabular}{c}
         \includegraphics[width=0.9\textwidth]{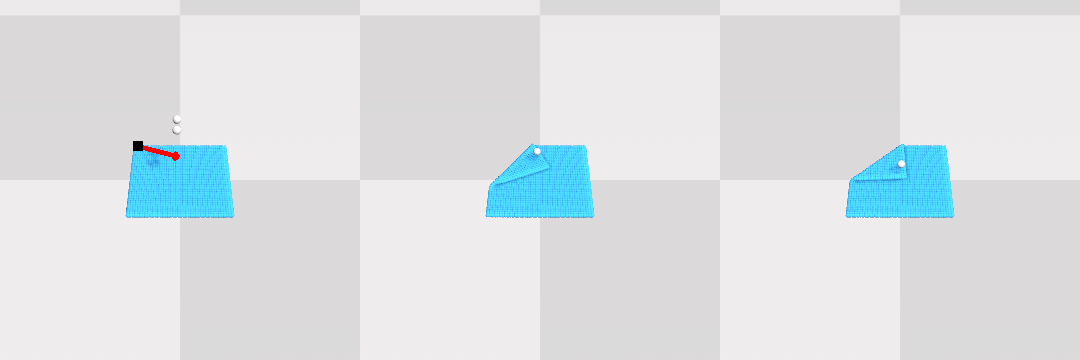}\\
    (a) Cost: groundtruth mapping \\
    \includegraphics[width=0.9\textwidth]{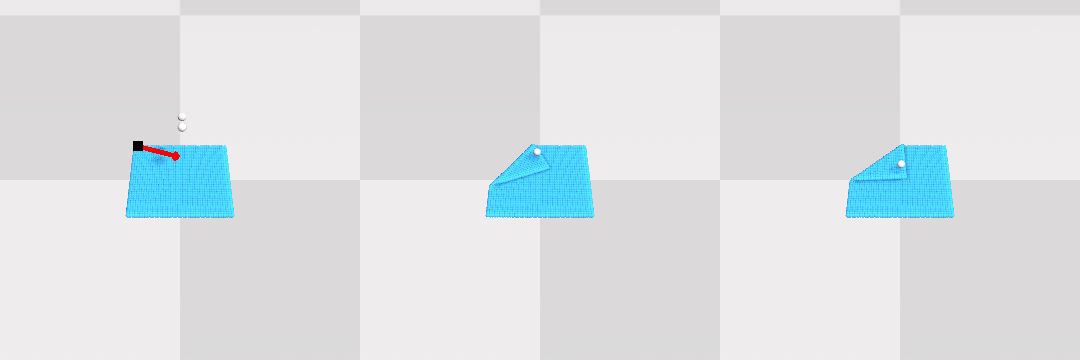} \\
        (b) Cost: Chamfer distance \\
    \includegraphics[width=0.9\textwidth]{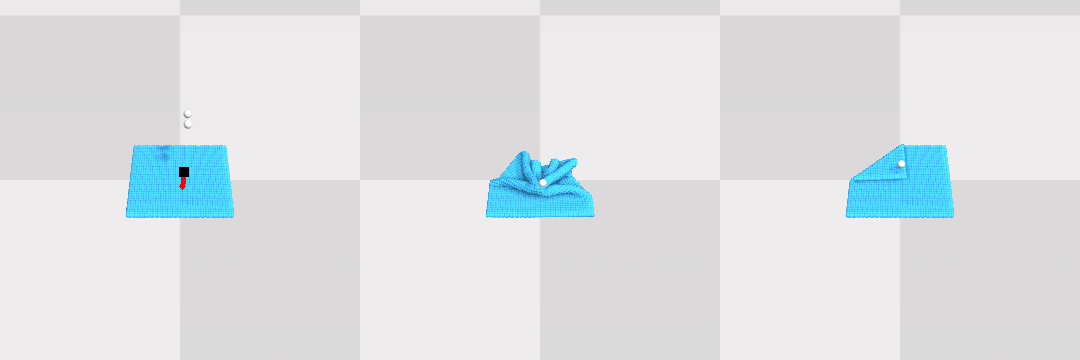} \\
    (c) Cost: IOU
    \end{tabular}

    \caption{VCD for folding, one-corner-in goal. The left column is the planned action, the middle column is the final achieved cloth state, and the right column is the goal.}
    \label{fig:folding1}
\end{figure}

\begin{figure}
    \centering
    \begin{tabular}{c}
         \includegraphics[width=0.9\textwidth]{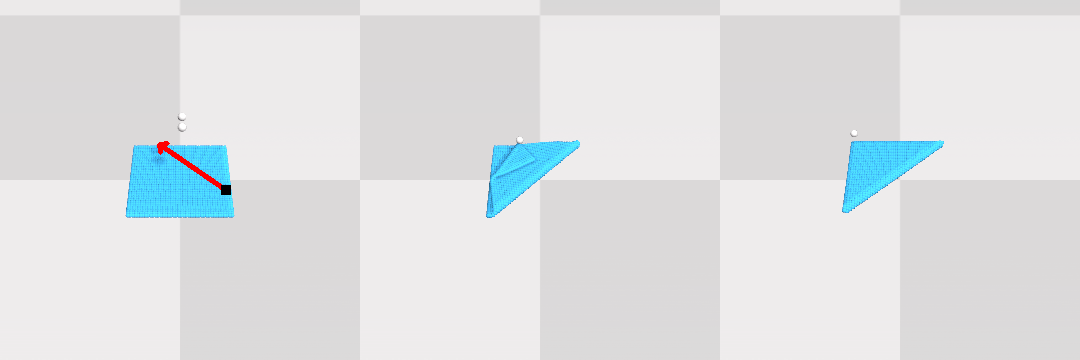}\\
    (a) Cost: groundtruth mapping \\
    \includegraphics[width=0.9\textwidth]{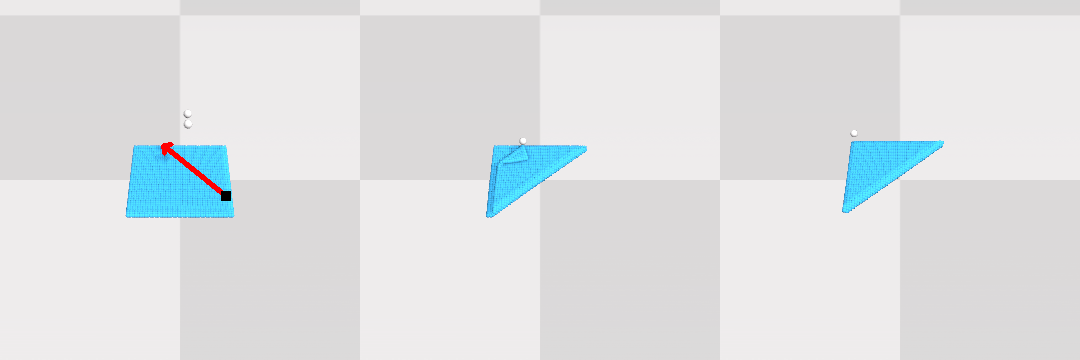} \\
        (b) Cost: Chamfer distance \\
    \includegraphics[width=0.9\textwidth]{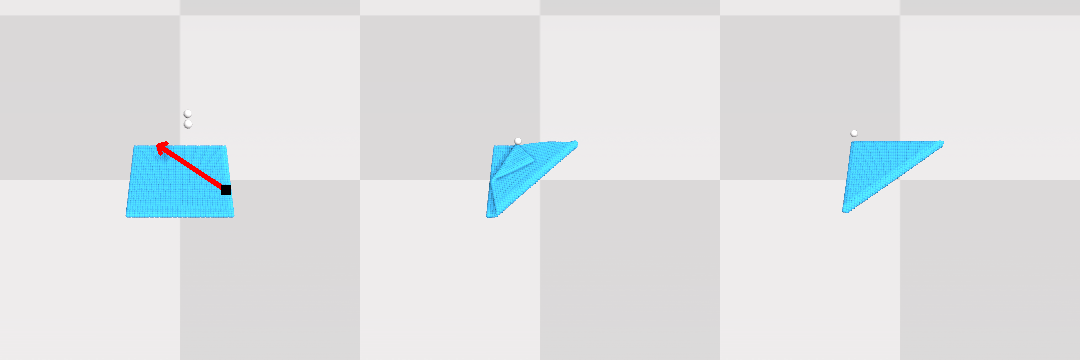} \\
    (c) Cost: IOU
    \end{tabular}

    \caption{VCD for folding, diagonal goal. The left column is the planned action, the middle column is the final achieved cloth state, and the right column is the goal.}
    \label{fig:folding2}
\end{figure}

\begin{figure}
    \centering
    \begin{tabular}{c}
         \includegraphics[width=0.9\textwidth]{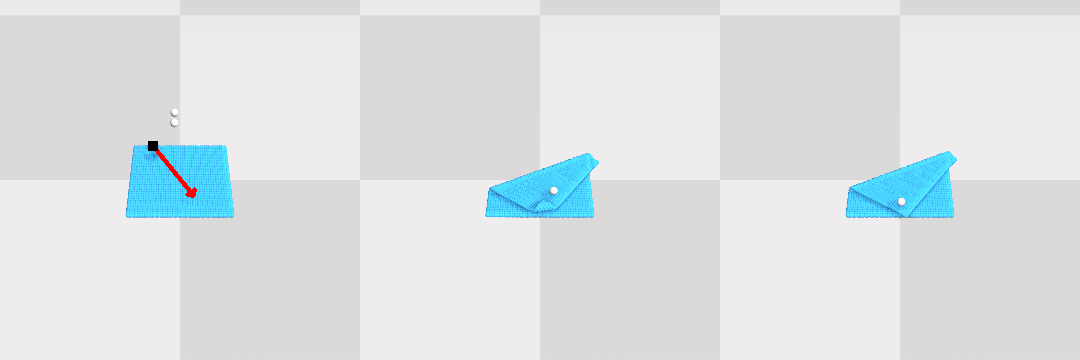}\\
    (a) Cost: groundtruth mapping \\
    \includegraphics[width=0.9\textwidth]{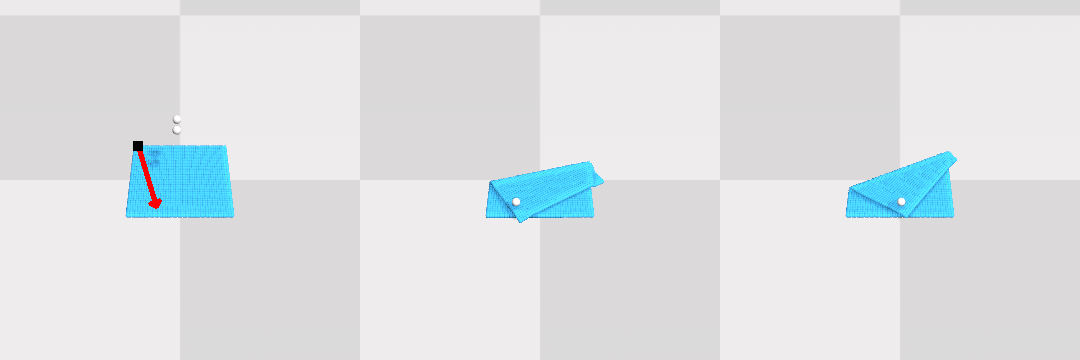} \\
        (b) Cost: Chamfer distance \\
    \includegraphics[width=0.9\textwidth]{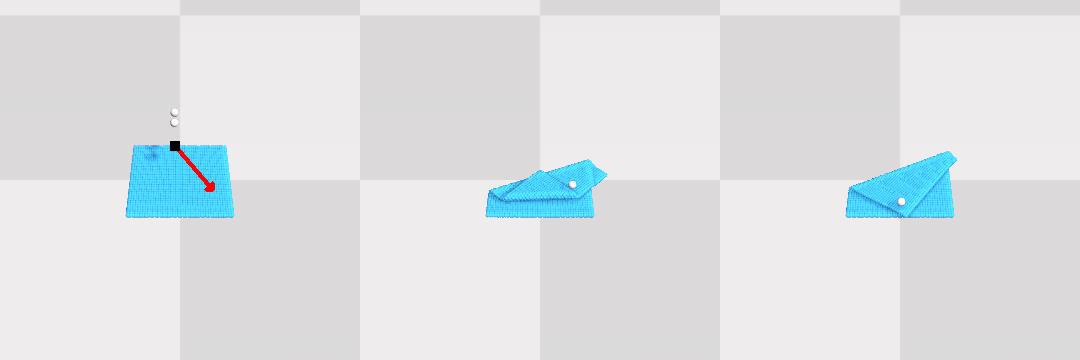} \\
    (c) Cost: IOU
    \end{tabular}

    \caption{VCD for folding, arbitrary goal. The left column is the planned action, the middle column is the final achieved cloth state, and the right column is the goal.}
    \label{fig:folding3}
\end{figure}

We show that VCD can also be used for cloth folding. We assume an initially flattened cloth is given, which can be obtained via planning with VCD for smoothing. Given a goal configuration of a target folded cloth (e.g., a diagonal fold for square cloth), we use VCD with CEM to plan actions that fold the cloth into the target configuration. We explore the following three different goal specifications and cost functions for the CEM planning:
\begin{itemize}
    \item A ground-truth cost function and goal specification that assumes access to the simulator cloth particles. The goal configuration of the cloth is specified as the goal locations of all particles. Given the voxelized point cloud of the initially flattened cloth, we first find a nearest neighbor mapping from each point in the point cloud to the simulator particles. The cost is then computed as the distance between the points in the achieved point cloud and their corresponding nearest-neighbor particles in the goal configuration. 
    \item We use the point cloud of the cloth for goal specification and Chamfer distance as the cost. Specifically, the cost is the Chamfer distance between the achieved point cloud and the goal point cloud.
    \item We use the depth image of the cloth for goal specification and 2D IOU as the cost. Specifically, we compute the intersection over union between the segmented achieved depth map and the segmented goal depth map as the cost.
\end{itemize}
We evaluate VCD on three goals as shown in \figurename~\ref{fig:folding1},~\ref{fig:folding2},~\ref{fig:folding3}: (1) one-corner-in, which folds one corner of the square cloth towards the center; (2) diagonal, which folds one corner of the square cloth towards the diagonal corner; (3) arbitrary, which folds one corner of the square cloth towards the middle point of the opposite edge.
For evaluation, we report the average particle distance between the achieved cloth state and the goal cloth state. The numerical results are shown in \tablename~\ref{tab:folding} and the qualitative results are shown in \figurename~\ref{fig:folding1},~\ref{fig:folding2},~\ref{fig:folding3}. 

As the result shows, VCD can be applied for folding with the above three ways for goal specification. For the ground-truth goal specification and cost computation using simulator particles, VCD performs fairly well for folding (average particle error within 0.3 - 1.3 cm, also see \figurename~\ref{fig:folding1},~\ref{fig:folding2},~\ref{fig:folding3} for qualitative results). With goal specification via point cloud and Chamfer distance as the cost, the performance of VCD is also reasonable (average particle error 0.3 - 2 cm, also see below for qualitative results), making it a practical choice to apply VCD for folding in the real world.

We also note that this VCD model is trained with random pick-and-place actions; the folding performance could be further improved if we add bias (such as corner grasping) during data collection to train VCD with more folding motions.

\begin{table}[h]
    \centering
    \begin{tabular}{c|c|c|c}
    \toprule
         & One-corner-in & Diagonal & Arbitrary  \\ \hline
    Ground-truth mapping  & 3.480 & 13.466 & 4.136 \\
    Chamfer distance     & 3.311 & 19.398  & 18.132 \\
    IOU &36.897 & 15.744 & 19.871 \\
    \bottomrule
    \end{tabular}
    \caption{Average particle distance (mm) between final achieved cloth state and goal cloth state.}
    \label{tab:folding}
\end{table}

\clearpage
\section{Robustness to Depth Sensor Noise}
\begin{figure}[h]
    \centering
    \includegraphics[width=0.4\textwidth]{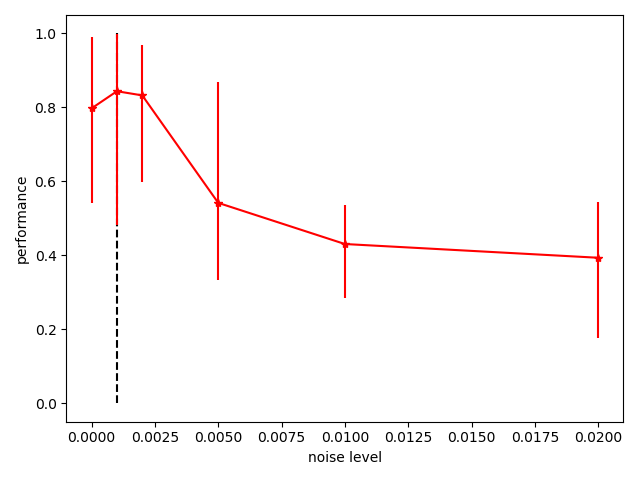}
    \caption{Normalized Improvement of VCD under different levels of depth sensor noise, with a maximal number of 10 pick-and-place actions for smoothing. The vertical dashed line represents the typical level of Azure Kinect noise, which is the depth sensor that we use for the real-world experiment. The error bars show the 25\% and 75\% percentile.}
    \label{fig:noise}
\end{figure}

When deployed in the real world, VCD might suffer from the depth camera noise. To investigate this, we manually add different levels of noise (Gaussian noise with different levels of variance) to the depth map in the simulation and test VCD's planning performance (with a maximal number of 10 pick-and-place actions). The result is shown in \figurename~\ref{fig:noise}. The dashed vertical line is the noise level of Azure Kinect depth camera that we use in the real world, as measured by Michal et al.~\cite{tolgyessy2021evaluation}. As shown, VCD is quite robust within the noise range of the Azure Kinect depth sensor.

\section{Comparison to Oracle using the  FleX Cloth Model}
\begin{figure}[h]
    \centering
    \includegraphics[width=0.75\textwidth]{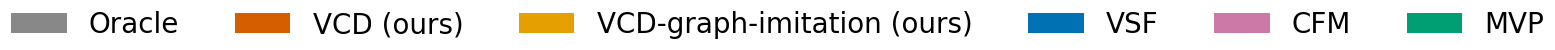}
    \includegraphics[width=.8\textwidth]{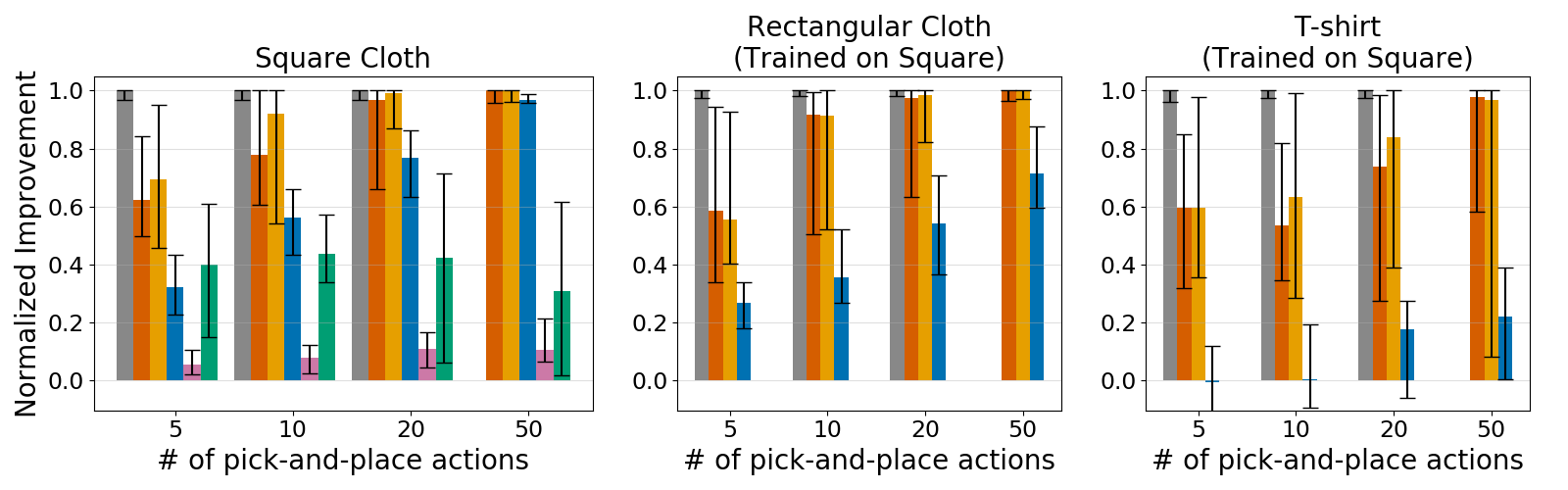}
    \caption{Normalized improvement on square cloth (left), rectangular cloth (middle), and t-shirt (right) for varying number of pick-and-place actions. The height of the bars show the median while the error bars show the 25 and 75 percentile.}
    \label{fig:oracle_result}
\end{figure}

How good can the system be if we know the full cloth dynamics? To answer this question, for our simulation experiments (shown in \figurename~\ref{fig:oracle_result}), we additionally show the performance of an oracle that uses the FleX cloth model for planning in \figurename~\ref{fig:oracle_result}. Here, oracle uses the same planning method as VCD and achieves perfect results in different clothes. This shows that better performance can be achieved if the full cloth model and dynamics can be better estimated, which we leave for future work.

\clearpage
\section{Ablations on architectural choices}
\begin{figure*}[h]
    \centering
    \includegraphics[width=1\textwidth]{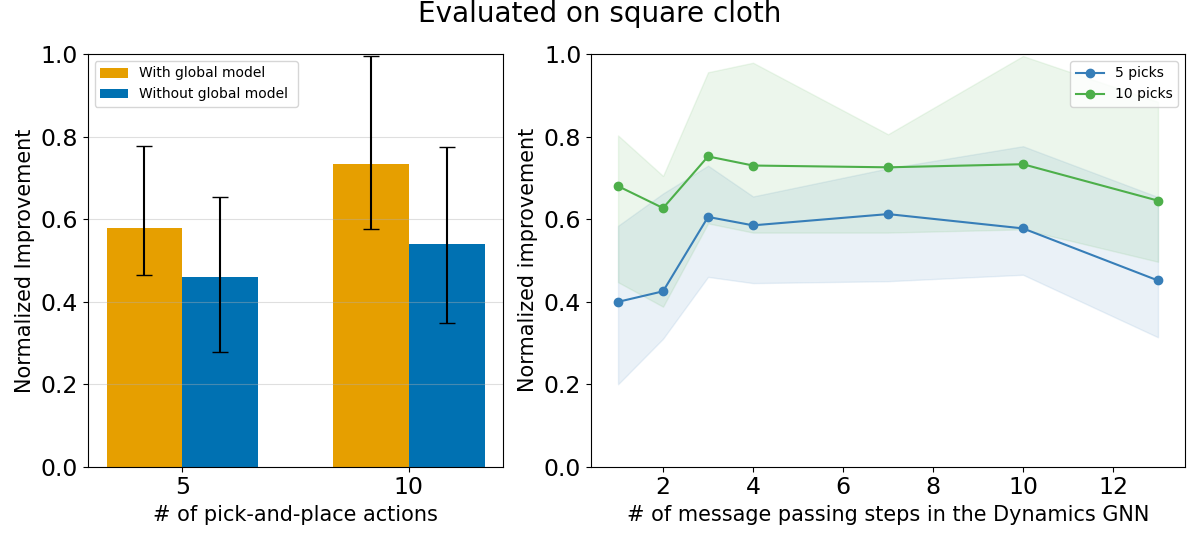}
    \caption{We evaluate the effects of a global model and the number of message passing steps in the dynamics GNN on the square cloth. The left figure shows that the usage of a global model is helpful to the planning performance. The right figure shows that our model is generally robust to the number of message passing steps as long as the number lies within the range of [3, 10].}
    \label{fig:archi_ablation}
\end{figure*}

For our edge and dynamics GNNs, we adopt the model architecture from GNS~\cite{sanchez2020learning}, as described in Appendix~\ref{sec:GNN Architecture}. In Sanchez-Gonzalez, et al~\cite{sanchez2020learning}, a comprehensive analysis on architectural design decisions for the GNS model was investigated.
We modify the GNS architecture by adding a global model in each GN block of the processor, which has the potential to speed up the propagation of information across the graph. The global model has been widely used in previous works in graph neural networks~\cite{battaglia2018relational,wang2020global, hamrick2018relational}.   
\figurename~\ref{fig:archi_ablation} (left) shows that using a global model in the dynamics model yield better planning performance than without it.  

We also evaluate the sensitivity of our dynamics model to the number of message passing steps ($L$). As shown in the right figure of \figurename~\ref{fig:archi_ablation}, our dynamics model is robust to a broad range of values for the number of message passing steps. We speculate that, when the number of message passing is too small, the effect of action cannot propagate to the particles that are distant from the picked point. With too many message passing steps, the model is prone to overfitting. Nonetheless, \figurename~\ref{fig:archi_ablation} (right) shows that there is a broad of values for the number of message passing steps that lead to similar performance; thus, our model is fairly robust to this parameter.

\end{document}